\documentclass[journal]{IEEEtai}

\usepackage[colorlinks,urlcolor=blue,linkcolor=blue,citecolor=blue]{hyperref}

\usepackage{color,array}

\usepackage{graphicx}

\usepackage[utf8]{inputenc} 
\usepackage[T1]{fontenc}    
\usepackage{hyperref}       
\usepackage{url}            
\usepackage{booktabs}       
\usepackage{amsfonts}       
\usepackage{nicefrac}       
\usepackage{microtype}      
\usepackage{lipsum}
\usepackage{fancyhdr}       
\usepackage{amsmath}
\usepackage{color}
\usepackage[table]{xcolor}

\usepackage{textcomp}
\usepackage{algorithm}
\usepackage{algpseudocode}
\usepackage{mathtools}
\usepackage{multirow}
\usepackage{longtable, booktabs}

\usepackage{array}
\usepackage{tabularx}
\usepackage{graphicx}       

\usepackage{subcaption}
\usepackage{pgffor}
\usepackage{booktabs,longtable,pdflscape}

\begin{document}

\title{Physics-Guided Generative AI for Property-Targeted 3D Porous Media Design}

\author{
Peng Wang
\thanks{
Peng Wang is with the Centre for Vision, Speech and Signal Processing (CVSSP),
University of Surrey, Guildford, UK.
}
}


\maketitle

\begin{abstract}
Inverse design of three-dimensional porous media is central to applications in filtration, catalysis, energy storage, fuel cells, thermal management, and biomedical scaffolds, but remains challenging because many distinct pore geometries can share similar porosity or permeability while small structural changes can strongly affect transport behaviour. This paper proposes a physics-guided generative AI framework for property-targeted porous media design, combining a property-aware variational autoencoder, a conditional latent diffusion model, and an independently trained differentiable structure-to-property surrogate. The framework learns a compact, physically informative latent design space, generates porous structures conditioned on target porosity and directional permeability, and refines generated samples using property-level feedback during denoising and decoding. Experiments on procedurally generated structures and real micro-CT porous-media datasets show improved target-property matching, directional permeability control, and property correlation compared with representative property-aware variational-autoencoder and latent-diffusion baselines. The results demonstrate a scalable route towards controllable inverse design of complex porous geometries and establish a foundation for simulation-informed generative AI tools in engineering and advanced materials discovery.
\end{abstract}

\begin{IEEEImpStatement}
Porous materials are essential in many technologies that support clean energy, sustainable manufacturing, healthcare, and environmental engineering. However, designing porous structures with desired physical properties remains slow and expensive because candidate designs often need to be evaluated through repeated numerical simulation or laboratory testing. This work introduces a physics-guided generative AI framework that can create three-dimensional porous structures conditioned on target porosity and directional permeability. By combining latent diffusion with differentiable physical-property feedback, the method improves agreement between generated designs and desired engineering properties while preserving realistic pore structures. The approach could reduce trial-and-error design cycles and support faster discovery of porous materials for applications such as battery electrodes, filtration membranes, fuel-cell components, thermal-management devices, biomedical scaffolds, and additively manufactured materials. The framework also provides a broader route towards trustworthy AI-assisted inverse design for complex geometry-to-property problems in engineering and physical sciences.
\end{IEEEImpStatement}


\section{Introduction}

Porous materials are widely used in applications including catalysis, filtration, energy storage, and carbon storage, where internal pore morphology directly governs effective physical properties such as porosity, permeability, and multiphase transport behaviour. This makes accurate reconstruction and controllable generation of three-dimensional (3D) porous structures critical for accelerating material analysis, simulation, and inverse design \cite{amiri2024new,zhu2025diffusion}. However, controllable porous-media generation remains fundamentally challenging because the physical behaviour of porous materials depends on complex internal geometries that are difficult to reconstruct and optimise efficiently. This has motivated growing interest in data-driven generative models capable of reconstructing realistic 3D porous structures from limited observations \cite{amiri2024new,naiff2025controlled}.

Early progress in this direction was driven largely by generative adversarial networks (GANs). Previous studies have shown that GAN-based approaches can generate realistic 3D porous structures either directly from volumetric training data or inferred from limited 2D observations. For instance, SliceGAN and related methods demonstrated the ability to reproduce structurally and morphologically meaningful porous geometries while alleviating imaging constraints \cite{amiri2024new,kench2022microlib,kononov2023reconstruction,volkhonskiy2022generative}. Subsequent studies further introduced property control by steering generation toward target properties or pore-scale statistics through latent-space optimisation, reinforcement learning, or physics-informed constraints \cite{ren2024using,nguyen2022synthesizing}. However, GAN-based models are still affected by factors such as unstable optimisation, mode collapse, and imperfect latent-space organisation, etc. \cite{dureth2023conditional,zhang2024vegan}.

More recently, diffusion models have emerged as a strong alternative for microstructure reconstruction and porous-media generation~\cite{nguyen2025camox}. Compared with GANs, diffusion models generally provide more stable optimisation and better coverage of complex data distributions. Existing studies have demonstrated that diffusion-based methods can generate visually realistic and statistically consistent porous structures while preserving important morphological characteristics \cite{dureth2023conditional}. In porous-media applications, diffusion-based approaches have further shown the ability to reproduce pore-space geometry that relate to physical properties, while latent diffusion formulations enable stable and larger-volume generation with reduced computational cost \cite{zhu2025diffusion,naiff2025controlled}. These developments suggest that diffusion models provide a promising foundation for controllable porous-media generation.

At the same time, inverse-design research has highlighted the importance of structured latent representations for linking porous geometry and effective physical properties~\cite{zhou2022two,klopries2025itf}. Property-aware variational autoencoders (pVAEs), a property-augmented variants of variational autoencoders, are one of the representative approaches that are exploited to map complex porous microstructures into compact and continuous latent spaces, to support structure reconstruction and property prediction \cite{wang2020deep,nguyen2026deep}. When combined with surrogate models for effective properties characterisation, such latent representations enable efficient exploration of structure-property relationships without repeated expensive numerical simulation and physical experiments \cite{nguyen2026deep,alzahrani2023pore,jones2024multiscale}. This is particularly attractive for porous media, where permeability evaluation through direct numerical simulation remains computationally expensive, while laboratory experimental evaluation is both time-consuming and financially costly.

Despite these advances, several important challenges remain. First, porous-media inverse design is inherently an ill-posed one-to-many mapping problem, where multiple structurally distinct porous geometries may correspond to similar effective properties. Second, existing controllable generation methods often lack differentiable closed-loop optimisation mechanisms capable of refining generated structures using physical property-aware feedback. Third, a mismatch may arise between diffusion-generated latent samples and the latent distribution originally learned by the porous-structure decoder (e.g., the pVAE decoder), reducing the physical consistency and controllability of generated porous structures during inverse-design.

\begin{figure*}[t]
    \centering
    \includegraphics[width=\textwidth]{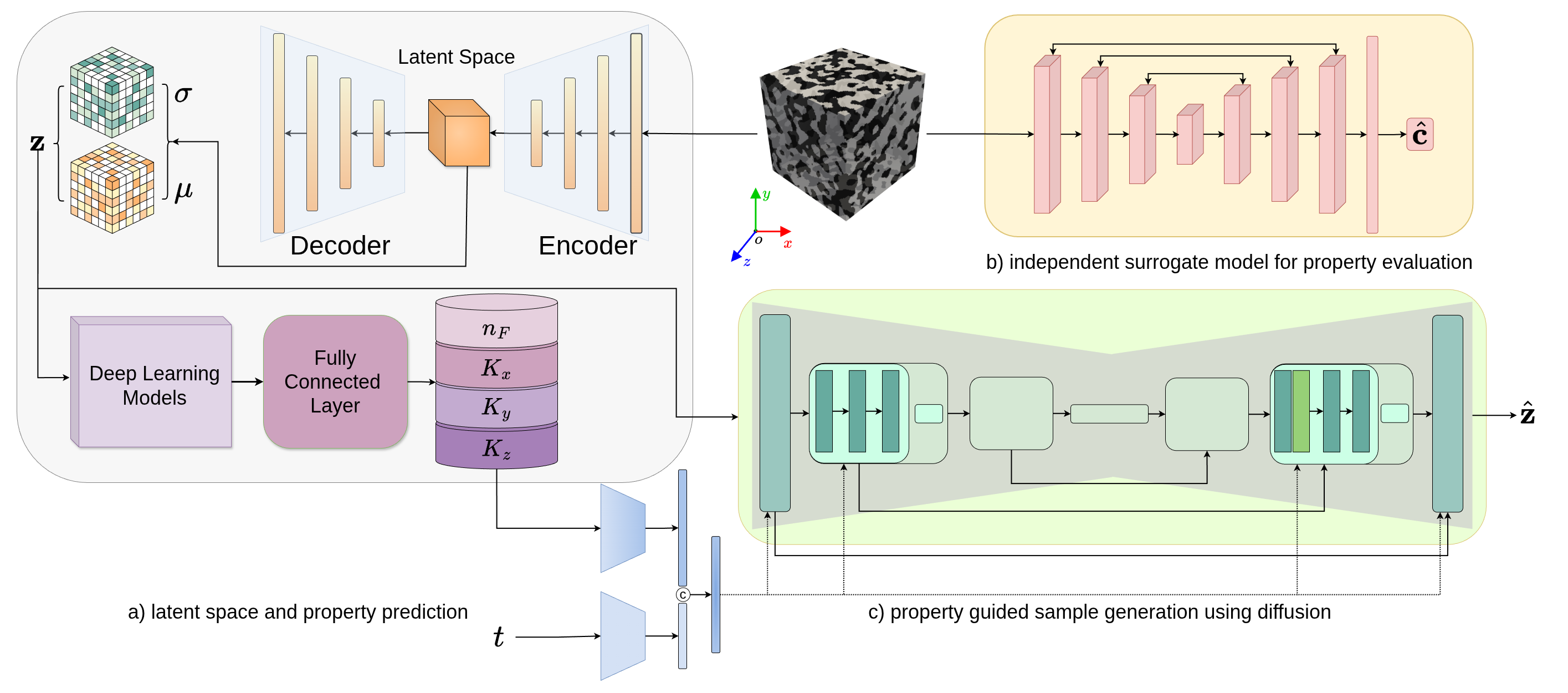}
    \caption{Overview of the proposed physical-property-guided latent diffusion framework for porous media inverse design. 
    (a) A property-aware variational autoencoder maps a 3D porous structure into a compact latent representation parameterised by $\boldsymbol{\mu}$ and $\boldsymbol{\sigma}$, while a latent property head encourages the representation to encode porosity $n_F$ and directional permeability $(K_x,K_y,K_z)$. 
    (b) An independently trained structure-to-property surrogate predicts the physical-property vector $\hat{\mathbf{c}}$ from generated porous structures and provides differentiable feedback during refinement. 
    (c) A conditional latent diffusion model uses the target property condition and timestep embedding to generate property-guided latent samples, which are decoded into candidate porous structures and subsequently verified using Palabos.}
    \label{fig:pipeline}
\end{figure*}

Motivated by these limitations, this work proposes a physical-property-guided latent diffusion framework for controllable porous-media inverse design. The proposed framework combines property-aware latent representation learning, conditional latent diffusion, and differentiable surrogate-based property evaluation within a unified optimisation pipeline. Generated porous structures are assessed using a differentiable structure-to-property predictor, enabling property discrepancies to be propagated back through the generation process to improve target-property consistency without requiring expensive online simulation or experimental feedback. The pipeline of the proposed work is shown in Fig.~\ref{fig:pipeline}. 

The main contributions of this work are summarised as follows:

\begin{itemize}

\item We propose a conditional latent diffusion framework for inverse porous-media design that integrates physical-property-aware latent representation learning with target-conditioned porous-structure generation.

\item We introduce a differentiable denoiser-decoder joint refinement mechanism that propagates surrogate-based property discrepancies through the generation pipeline, improving compatibility between diffusion-generated latent samples and physically meaningful porous geometries.

\item We develop a physics platform-verified porous-media generation framework that combines pVAE latent modelling, conditional latent diffusion, differentiable structure-to-property prediction, and Palabos-based permeability evaluation within a unified optimisation pipeline.

\item We perform extensive experiments on synthetic and real micro-CT porous-media datasets, demonstrating improved permeability controllability, target-property consistency, and simulator-verified inverse-design performance compared with representative pVAE-based and latent-diffusion-based baselines.

\end{itemize}

\section{Related Work}

Inverse porous-material design requires simultaneously addressing two tightly coupled challenges: controllable porous-structure generation and reliable physical-property evaluation. The former is inherently an ill-posed one-to-many mapping problem, where multiple distinct porous geometries may correspond to similar effective properties. The latter is computationally challenging because accurate permeability and transport evaluation typically relies on expensive numerical simulation or physical experiments. Existing works have therefore focused primarily on either generative porous reconstruction or physics-based property characterisation, while relatively few studies investigate differentiable closed-loop inverse-design frameworks that integrate both generation and property-aware optimisation.

\subsection{Property-Guided Generation for Inverse Material Design}

Most existing porous-media generation works primarily follow a \emph{forward design} paradigm, where 3D porous structures are reconstructed from 2D micro-CT observations and subsequently evaluated through physics-based simulation. Recent diffusion-based methods have demonstrated improved stability and generation quality for porous-media reconstruction. Zhu \emph{et al.} showed that diffusion models can effectively reproduce pore morphology and porosity distributions that are critical for downstream physical properties \cite{zhu2025diffusion}. Naiff \emph{et al.} further introduced controlled latent diffusion models for porous-media reconstruction, enabling larger-volume generation and improved coverage of porous-media statistics \cite{naiff2025controlled}. However, these methods primarily focus on controllable reconstruction, rather than differentiable inverse-design refinement guided by simulator-consistent property feedback.

Beyond reconstruction, many practical applications require generated porous structures to satisfy prescribed physical properties. This has led to increasing interest in inverse design and property-guided synthesis. Early inverse-design approaches primarily relied on GAN-based frameworks. For example, Nguyen \emph{et al.} proposed a combined GAN and actor-critic reinforcement learning framework for synthesizing porous microstructures with controllable structural properties \cite{nguyen2022synthesizing}. Ren and Srinivasan similarly investigated property-constrained porous generation through latent-space deformation guided by pore-network-derived physical attributes~\cite{ren2024using}.

More recently, VAE-based latent modelling has emerged as a more structured approach for porous-media inverse design. Nguyen \emph{et al.} developed a pVAE-based inverse-design framework in which latent representations are jointly supervised by reconstruction and physical-property prediction \cite{nguyen2026deep}. By combining latent modelling with surrogate-based permeability prediction, the framework enables efficient structure-property exploration and gradient-based optimisation. This property-aware latent-space paradigm is particularly relevant because it provides a continuous and interpretable latent representation that is more suitable for controllable generation and inverse design than conventional GAN latent codes. Because recent porous-media inverse-design research has increasingly shifted toward latent diffusion and property-aware latent modelling, this work primarily compares against representative pVAE-based and latent-diffusion-based approaches that are more closely aligned with the proposed framework.

\subsection{Physical Simulation for Real-time Property Evaluation}

Generative inverse-design frameworks ultimately depend on reliable structure-property datasets, which in turn require accurate physics-based simulation or experimental characterisation. In porous-media inverse design, real-time evaluation of generated structures is particularly important because property discrepancies can potentially be used to guide the generation process through differentiable optimisation.

Phu \emph{et al.} investigated deformation-dependent permeability estimation using a pipeline that combines nano-CT reconstruction with Lattice Boltzmann Method (LBM) simulation in Palabos \cite{phu2024investigating}. Their work demonstrated how permeability tensors can be estimated from realistic 3D porous geometries through physics-based simulation. Lavigne \emph{et al.} proposed an open-source framework for synthetic porous-microstructure generation and permeability analysis by integrating geometric construction, meshing, and fluid-structure interaction simulation \cite{lavigne2025synthetic}. Although these works are not themselves generative-learning frameworks, they demonstrate the importance of physics-based simulation for reliable porous-media property evaluation.

In parallel, indepent and reliable learning-based surrogate models are gaining attention for accelerating permeability prediction and structure-property analysis. Compared with repeated numerical simulation, surrogate models can provide significantly faster property estimation while remaining differentiable, making them attractive for integration within generative inverse-design pipelines. However, relatively few existing works combine latent diffusion, differentiable surrogate evaluation, and simulator-verified inverse-design refinement within a unified porous-media generation framework.

\section{Methodology}
\label{sec:method}

\subsection{Latent Diffusion Inverse Design Framework}
\label{subsec:method_framework}

Let $\mathbf{x}\in\{0,1\}^{w\times h\times l}$ denote a binary porous microstructure of width $w$, height $h$, and depth $l$, where $1$ represents the pore phase and $0$ represents the solid phase. Each sample is associated with a raw physical-property vector. In this work, we focus on pore fraction and directional permeability, and define
\begin{equation}
    \mathbf{y}
    =
    [n_F,K_x,K_y,K_z]^T,
    \label{eq:property_vector}
\end{equation}
where $n_F$ is the pore fraction, and $K_x$, $K_y$, and $K_z$ are directional permeabilities along the three principal axes. It is worth noting that permeability values typically span several orders of magnitude and are often more naturally interpreted through relative rather than absolute differences. Training directly on raw permeability values can cause large-permeability samples to dominate the regression loss, leading to unstable optimisation and poorer accuracy in low-permeability regimes. Therefore, for stable model training, we have converted permeability through
\begin{equation}
\mathbf{c} =
\begin{bmatrix}
    (n_F-\mu_{n_F})/\sigma_{n_F} \\[2pt]
    (\log K_x-\mu_{\log K_x})/\sigma_{\log K_x} \\[2pt]
    (\log K_y-\mu_{\log K_y})/\sigma_{\log K_y} \\[2pt]
    (\log K_z-\mu_{\log K_z})/\sigma_{\log K_z}
\end{bmatrix}.
\label{eq:normalized_condition}
\end{equation}
We compute all property-based losses in the normalised/log-property space. But physical property verification and reported physical errors are computed after inverting Eq.~\eqref{eq:normalized_condition} back to raw physical units.

The goal of inverse porous-media design is to generate a porous structure $\mathbf{x}_{\mathrm{gen}}$ whose physical properties match a raw target property vector $\mathbf{y}^{*}$. Since the models operate in the normalised condition space, this can be written as learning a conditional generative model
\begin{equation}
    G_{\theta}:(\mathbf{c}^{*},\boldsymbol{\xi})\mapsto \mathbf{x}_{\mathrm{gen}},
    \label{eq:conditional_generator}
\end{equation}
where $\mathbf{c}^{*}$ is obtained by normalising the raw target vector $\mathbf{y}^{*}$ using Eq.~\eqref{eq:normalized_condition}, $\boldsymbol{\xi}$ denotes stochastic latent noise, and $\theta$ denotes model parameters.

The physical properties of a generated structure can be evaluated by high-fidelity LBM-based simulators such as Palabos:
\begin{equation}
    \mathbf{y}_{\mathrm{sim}} = S(\mathbf{x}_{\mathrm{gen}}),
    \label{eq:simulator_mapping}
\end{equation}
where $S(\cdot)$ denotes the LBM-based simulator. However, $S(\cdot)$ is computationally expensive and non-differentiable with respect to the generator parameters. Therefore, during training and refinement, we use a differentiable surrogate model
\begin{equation}
    \hat{\mathbf{c}} = s_{\psi}(\mathbf{x}_{\mathrm{gen}}),
    \label{eq:surrogate_mapping_method}
\end{equation}
where $s_{\psi}$ is a deep learning based structure-to-property predictor trained to approximate the mapping from porous geometry to normalised physical properties.

The framework consists of four components. First, a pVAE maps voxelised porous structures into property-aware compact latent representations and reconstructs them back to voxel space. Second, a conditional latent diffusion model learns the distribution of pVAE latents conditioned on target properties, and the learned distributions are decoded by the pVAE decoder to achieve porous media generation. This constitutes the generative model $G_\theta$. Third, an independently trained surrogate model that predicts $\mathbf{c}$ from voxel fields and provides differentiable property feedback to $G_\theta$ during denoiser-decoder refinement. Fourth, generated candidates are exported to Palabos for physics-based verification in raw physical units.

\subsection{Differentiable Property Feedback}
\label{subsec:differentiable_feedback}

Given a target condition $\mathbf{c}^{*}$, the generated structure is evaluated by the frozen surrogate:
\begin{equation}
    \hat{\mathbf{c}}
    =
    s_{\psi}(\mathbf{x}_{\mathrm{gen}}).
\end{equation}
The property-matching loss is
\begin{equation}
    \mathcal{L}_{\mathrm{prop}}
    =
    \left\|
    s_{\psi}(\mathbf{x}_{\mathrm{gen}})
    -
    \mathbf{c}^{*}
    \right\|_2^2 .
    \label{eq:prop_loss_general}
\end{equation}
Because both the generator and surrogate are differentiable neural networks, gradients can be propagated from the property discrepancy back through the generated voxel field:
\begin{equation}
    \frac{\partial \mathcal{L}_{\mathrm{prop}}}{\partial \theta}
    =
    \frac{\partial \mathcal{L}_{\mathrm{prop}}}{\partial \hat{\mathbf{c}}}
    \frac{\partial \hat{\mathbf{c}}}{\partial \mathbf{x}_{\mathrm{gen}}}
    \frac{\partial \mathbf{x}_{\mathrm{gen}}}{\partial \theta}.
    \label{eq:property_chain_rule}
\end{equation}
This enables property-guided refinement without requiring online LBM simulation or physical experiments during optimisation.

\subsection{Network Architectures}
\label{subsec:architecture_details}

\paragraph{Property-aware VAE.}
The pVAE operates on voxelised porous structures and maps each input $\mathbf{x}$ into a compact latent representation. The encoder is implemented using a residual convolutional backbone that progressively downsamples the input volume and extracts hierarchical geometry features. Two convolutional heads then predict the mean and log-variance of the approximate posterior,
\begin{equation}
q_{\phi}(\mathbf{z}\mid\mathbf{x})
=
\mathcal{N}
\left(
\boldsymbol{\mu}_{\phi}(\mathbf{x}),
\mathrm{diag}(\boldsymbol{\sigma}_{\phi}^{2}(\mathbf{x}))
\right).
\label{eq:pvae_posterior}
\end{equation}
Latent samples are obtained using the reparameterisation trick:
\begin{equation}
    \mathbf{z}
    =
    \boldsymbol{\mu}_\phi(\mathbf{x})
    +
    \boldsymbol{\sigma}_\phi(\mathbf{x})\odot\boldsymbol{\epsilon},
    \qquad
    \boldsymbol{\epsilon}\sim\mathcal{N}(\mathbf{0},\mathbf{I}).
    \label{eq:reparameterisation_method}
\end{equation}
The decoder maps the latent representation back to voxel space using residual upsampling blocks and produces a soft occupancy field $\hat{\mathbf{x}}\in[0,1]^{w\times h\times l}$ through a sigmoid output layer.

To make the latent space physically informative, the pVAE also includes a latent property head $h_\phi(\cdot)$ that predicts the normalised condition vector $\mathbf{c}$ from the latent representation. The pVAE objective is
\begin{equation}
\begin{aligned}
\mathcal{L}_{\mathrm{pVAE}}
=&\;
\lambda_{\mathrm{rec}}\mathcal{L}_{\mathrm{BCE}}
+
\lambda_{\mathrm{KL}}
D_{\mathrm{KL}}
\left(
q_\phi(\mathbf{z}\mid\mathbf{x})
\,\|\, 
\mathcal{N}(\mathbf{0},\mathbf{I})
\right) \\
&+
\lambda_{\mathrm{lat}}
\left\|
h_\phi(\mathbf{z})-\mathbf{c}
\right\|_2^2 .
\end{aligned}
\label{eq:pvae_loss}
\end{equation}
Here, $\mathcal{L}_{\mathrm{BCE}}$ is the binary cross entropy (BCE) that preserves voxel-level reconstruction fidelity, the KL term is the Kullback–Leibler divergence that regularises the latent distribution to remain smooth and sampleable, and the latent-property loss encourages the latent representation to encode physically meaningful information.

\paragraph{Conditional latent diffusion denoiser.}
The conditional diffusion model operates in the pVAE latent space rather than directly in the high-dimensional voxel space for stability~\cite{rombach2022high}. Let $\mathbf{z}_0$ denote a latent sample obtained from the pVAE encoder. The forward diffusion process gradually corrupts $\mathbf{z}_0$ with Gaussian noise:
\begin{equation}
    \mathbf{z}_t
    =
    \sqrt{\bar{\alpha}_t}\mathbf{z}_0
    +
    \sqrt{1-\bar{\alpha}_t}\boldsymbol{\epsilon},
    \qquad
    \boldsymbol{\epsilon}\sim\mathcal{N}(\mathbf{0},\mathbf{I}).
    \label{eq:forward_diffusion_method}
\end{equation}
A residual U-Net denoiser $\boldsymbol{\epsilon}_{\theta}$ is trained to predict the injected noise conditioned on the timestep $t$ and the normalised property condition $\mathbf{c}$:
\begin{equation}
    \mathcal{L}_{\mathrm{diff}}
    =
    \mathbb{E}_{\mathbf{z}_0,t,\boldsymbol{\epsilon}}
    \left[
    \left\|
    \boldsymbol{\epsilon}
    -
    \boldsymbol{\epsilon}_{\theta}(\mathbf{z}_t,t,\mathbf{c})
    \right\|_2^2
    \right].
    \label{eq:diffusion_loss}
\end{equation}

Conditioning is injected through FiLM residual blocks. A sinusoidal timestep embedding and an embedding of the property condition $\mathbf{c}$ are combined and used to modulate intermediate feature maps:
\begin{equation}
    \mathrm{FiLM}(\mathbf{h},\mathbf{c},t)
    =
    \mathbf{h}\odot(1+\gamma(\mathbf{c},t))
    +
    \beta(\mathbf{c},t),
    \label{eq:film_conditioning}
\end{equation}
where $\mathbf{h}$ denotes an intermediate feature map in the denoiser, while $\gamma(\mathbf{c},t)$ and $\beta(\mathbf{c},t)$ are feature-wise scale and shift parameters predicted from the joint condition embedding. In this work, $\mathbf{z}_0$ is sampled from the pVAE posterior rather than using only the encoder mean, so that diffusion training better matches the latent distribution consumed by the decoder during pVAE training.

At inference time, the conditional generator starts from a sampled latent noise tensor $\mathbf{z}_T\sim\mathcal{N}(\mathbf{0},\mathbf{I})$ and applies the learned reverse denoising process conditioned on $\mathbf{c}$ to obtain a generated latent sample $\hat{\mathbf{z}}_0$, which is then decoded by the pVAE decoder.

\paragraph{Structure-to-property surrogate.}
The surrogate $s_\psi(\mathbf{x})$ is an independently trained residual convolutional network that predicts the normalised/log-property vector from a voxelised porous structure:
\begin{equation}
    \hat{\mathbf{c}} = s_\psi(\mathbf{x}).
    \label{eq:surrogate_prediction_method}
\end{equation}
It is trained using
\begin{equation}
    \mathcal{L}_{\mathrm{sur}}
    =
    \left\|
    s_\psi(\mathbf{x})-\mathbf{c}
    \right\|_2^2.
    \label{eq:surrogate_loss}
\end{equation}
The surrogate is trained independently from the generative models and is frozen during all optimisation and refinement experiments. This ensures that property-guided improvements arise from updating the generator rather than from changing the evaluator.

\begin{algorithm}[t]
\caption{Training and physics-verified inverse design pipeline}
\label{alg:training_pipeline}
\begin{algorithmic}[1]
\Require Training geometries $\{\mathbf{x}_i\}$, raw properties $\{\mathbf{y}_i\}$, normalisation statistics, Palabos solver
\State Convert raw properties $\mathbf{y}_i=[n_F,K_x,K_y,K_z]^T$ to normalised conditions $\mathbf{c}_i$ using Eq.~\eqref{eq:normalized_condition}
\State Train the pVAE using Eq.~\eqref{eq:pvae_loss}
\State Train the surrogate model $s_\psi(\mathbf{x})$ to predict $\mathbf{c}_i$ from $\mathbf{x}_i$
\State Freeze the pVAE encoder, pVAE property head, and surrogate
\State Encode training samples, sample $\mathbf{z}_0$ from the pVAE posterior, and train the conditional latent denoiser using Eq.~\eqref{eq:diffusion_loss}
\State Initialise surrogate-guided refinement from the trained pVAE, surrogate, and denoiser checkpoints
\State In denoiser-decoder refinement mode, update only the denoiser and pVAE decoder
\For{each minibatch $(\mathbf{x},\mathbf{c})$}
    \State Encode $\mathbf{x}$ to sampled latent $\mathbf{z}_0$ using the frozen pVAE encoder
    \State Sample timestep $t$ and noise $\boldsymbol{\epsilon}$, and construct $\mathbf{z}_t$
    \State Predict $\hat{\boldsymbol{\epsilon}}_\theta(\mathbf{z}_t,t,\mathbf{c})$ and compute $\mathcal{L}_{\mathrm{diff}}$
    \State Estimate $\hat{\mathbf{z}}_0=(\mathbf{z}_t-\sqrt{1-\bar{\alpha}_t}\hat{\boldsymbol{\epsilon}}_\theta)/\sqrt{\bar{\alpha}_t}$
    \State Decode $\hat{\mathbf{z}}_0$ to $\hat{\mathbf{x}}_{\mathrm{gen}}$ and predict $\hat{\mathbf{c}}=s_\psi(\hat{\mathbf{x}}_{\mathrm{gen}})$
    \State Compute $\mathcal{L}_{\mathrm{prop}}=\|\hat{\mathbf{c}}-\mathbf{c}\|_2^2$
    \State Decode $\mathbf{z}_0$ to $\hat{\mathbf{x}}_{\mathrm{rec}}$ and compute $\mathcal{L}_{\mathrm{rec}}=\mathrm{BCE}(\hat{\mathbf{x}}_{\mathrm{rec}},\mathbf{x})$
    \State Optionally compute $\mathcal{L}_{\mathrm{anchor}}$ using Eq.~\eqref{eq:anchor_loss}
    \State Update trainable parameters with Eq.~\eqref{eq:finetune_loss}
\EndFor
\State For each target $\mathbf{y}^{*}$, convert it to $\mathbf{c}^{*}$, sample candidate latents, decode candidates, evaluate them with the surrogate, export selected geometries, and verify them with Palabos
\end{algorithmic}
\end{algorithm}

\subsection{Training and Refinement}
\label{subsec:training_algorithm}

The trained conditional diffusion module generates target-conditioned latent representations that are decoded by the pVAE decoder into voxelised porous structures for property evaluation. Although the diffusion model is trained on latent samples obtained from the pVAE encoder, the distribution of diffusion-generated latents may still differ from the posterior latent distribution on which the decoder was originally trained. This denoiser-decoder mismatch can lead to decoded structures whose surrogate-predicted or Palabos-verified properties deviate from the prescribed target condition.

To reduce this mismatch, we introduce a surrogate-guided refinement stage. Optimising only the diffusion denoiser can improve the generated latent samples, while optimising only the decoder can adapt the latent-to-voxel mapping. However, neither option fully captures the coupling between the denoising generated latent representation and the final decoded voxel structure. Therefore, this work proposes to refine the denoiser and decoder together. This allows the generated latent distribution and the reconstruction mapping to adapt jointly under the frozen surrogate property constraint, to improve overall generation performance.

During refinement, the pVAE encoder, pVAE latent property head, and structure-to-property surrogate are frozen. The refinement objective is
\begin{equation}
    \mathcal{L}_{\mathrm{ft}}
    =
    \lambda_{\mathrm{diff}}\mathcal{L}_{\mathrm{diff}}
    +
    \lambda_{\mathrm{prop}}\mathcal{L}_{\mathrm{prop}}
    +
    \lambda_{\mathrm{rec}}\mathcal{L}_{\mathrm{rec}}
    +
    \lambda_{\mathrm{anchor}}\mathcal{L}_{\mathrm{anchor}} .
    \label{eq:finetune_loss}
\end{equation}
The diffusion loss $\mathcal{L}_{\mathrm{diff}}$ is the same noise-prediction objective used for conditional latent diffusion training. It preserves the denoising behaviour while allowing the denoiser to adapt to the property-guided refinement objective. The property loss is computed on generated samples. Given a target condition $\mathbf{c}^{*}$, the refined denoiser produces a generated latent sample $\hat{\mathbf{z}}_0$, which is decoded into a voxel structure by the trainable decoder. The frozen surrogate then predicts its normalised/log-property vector:
\begin{equation}
    \mathcal{L}_{\mathrm{prop}}
    =
    \left\|
    s_\psi(D_\phi(\hat{\mathbf{z}}_0))-\mathbf{c}^{*}
    \right\|_2^2 .
    \label{eq:finetune_prop_loss}
\end{equation}
This term propagates the property discrepancy through the frozen surrogate and decoded voxel field back to the trainable generator components, enabling differentiable property-guided refinement without online Palabos simulation.

The reconstruction-preservation loss $\mathcal{L}_{\mathrm{rec}}$ is computed from real training samples. The frozen encoder maps an input structure $\mathbf{x}$ to a sampled latent representation $\mathbf{z}_0$, and the trainable decoder reconstructs the structure as $D_\phi(\mathbf{z}_0)$. The reconstruction is compared with $\mathbf{x}$ using binary cross-entropy:
\begin{equation}
    \mathcal{L}_{\mathrm{rec}}
    =
    \mathrm{BCE}(D_\phi(\mathbf{z}_0),\mathbf{x}) .
    \label{eq:finetune_rec_loss}
\end{equation}
This term helps preserve the decoder's ability to reconstruct valid porous structures from pVAE latents and prevents refinement from degrading the learned voxel manifold.

The optional anchor loss constrains the refined decoder to remain close to the original decoder:
\begin{equation}
    \mathcal{L}_{\mathrm{anchor}}
    =
    \left\|
    D_\phi(\mathbf{z}_0)-D_{\phi_0}(\mathbf{z}_0)
    \right\|_2^2 ,
    \label{eq:anchor_loss}
\end{equation}
where $D_{\phi_0}$ denotes the original frozen decoder before refinement. This term regularises decoder updates by discouraging large deviations from the original reconstruction mapping.

\subsection{Inference and Physical Platform Evaluation}
\label{subsec:inference}

Given a raw target property vector $\mathbf{y}^{*}$, the target is first converted to the normalised/log-property condition $\mathbf{c}^{*}$ using Eq.~\eqref{eq:normalized_condition}. The pVAE baseline samples a number of latent representations (e.g., 50) from the prior and decodes them using the frozen pVAE decoder. Latent space optimsation will be applied for pVAE baseline as was done in~\cite{nguyen2026deep}. The latent diffusion baseline samples a target-conditioned latent through the conditional reverse denoising process and decodes it using the original pVAE decoder. The proposed method samples a target-conditioned latent using the refined diffusion denoiser and decodes it using the refined decoder.

The generated candidate is evaluated by the surrogate directly after sampling and decoding. For each sample, we define the relative pore-fraction error as
\begin{equation}
e_{n_F}^{\mathrm{sur}}
=
\frac{
|\hat{n}_F-n_F^{*}|
}{
|n_F^{*}|+\delta
},
\end{equation}
and the relative permeability error as
\begin{equation}
e_{K_i}^{\mathrm{sur}}
=
\frac{
|\hat{K}_i-K_i^{*}|
}{
|K_i^{*}|+\delta
},
\qquad
i\in\{x,y,z\}.
\end{equation}
The raw relative evaluation score is then computed as
\begin{equation}
r(\hat{\mathbf{y}},\mathbf{y}^{*})
=
\sqrt{
(e_{n_F}^{\mathrm{sur}})^2
+
(e_{K_x}^{\mathrm{sur}})^2
+
(e_{K_y}^{\mathrm{sur}})^2
+
(e_{K_z}^{\mathrm{sur}})^2
}.
\label{eq:raw_rank_score}
\end{equation}
where $\delta$ is a small numerical constant used for numerical stability.

The generated samples in the voxel field is thresholded into a binary pore-solid structure and exported to Palabos for LBM-based verification. Final reported errors are computed from Palabos-measured properties in raw physical units, but these surrogate metrics are critical to enable real-time evaluations of the generated samples.

\section{Experiments}
\label{sec:experiments}

\subsection{Datasets}
\label{sec:exp:datasets}

Experiments are conducted on two datasets: a procedurally generated synthetic porous-media dataset and a real micro-CT porous-media dataset. In both datasets, each sample is represented as a binary voxel volume $\mathbf{x}\in\{0,1\}^{100\times100\times100}$, where $1$ denotes pore and $0$ denotes solid. Each sample is associated with the raw property vector defined in Eq.~\eqref{eq:property_vector}. The properties of interest are pore fraction $n_F$ and directional permeability $(K_x,K_y,K_z)$.

\paragraph{Synthetic dataset.}
The synthetic dataset is generated procedurally and labelled using the Palabos LBM permeability solver. Each voxelised structure is exported into a Palabos-compatible geometry file. Permeability is simulated independently along the $x$, $y$, and $z$ directions to obtain $(K_x,K_y,K_z)$. The synthetic dataset contains 17,000 samples.

\paragraph{Real micro-CT dataset.}
The real dataset is constructed from micro-CT scans. The original grayscale volumes are centrally cropped to remove scan boundaries and surrounding background regions. The cropped volumes are segmented into binary pore/solid structures and smoothed using a signed-distance-field-based operation to suppress isolated voxel noise while preserving pore connectivity. The processed volumes are divided into overlapping $100\times100\times100$ sub-volumes using a stride of 25 voxels. Each sub-volume is exported to Palabos and simulated along the three principal directions to obtain directional permeability labels. The real micro-CT dataset contains 5,525 samples.

\paragraph{Property normalisation.}
All models are trained using the normalised/log-property condition vector $\mathbf{c}$ defined in Eq.~\eqref{eq:normalized_condition}. Permeability values are log-transformed before standardisation, while pore fraction is standardised directly.

\paragraph{Dataset split.}
For both synthetic and real micro-CT datasets, 70\% of the samples are used for training, 10\% for validation, and 20\% for testing.

\subsection{Implementation Details}
\label{sec:exp:implementation}

For all reported experiments, the pVAE encodes each volume into a latent representation of size $32\times5\times5\times5$, with 32 the batch size and $5\times5\times5$ the bottleneck dimensions. More results can be found in Sec.~\ref{sec:ablation}. For the denoiser-decoder refinement setting,  we use $\lambda_{\mathrm{diff}}=1.0$, $\lambda_{\mathrm{prop}}=0.1$, $\lambda_{\mathrm{rec}}=0.1$, and $\lambda_{\mathrm{anchor}}=0.001$. 
The denoiser and decoder are updated with learning rates of $10^{-5}$ and $10^{-6}$, respectively. 

We will first report results without (or with minimal) latent space optimisation, and then report results with latent space optimisation in the ablation. This is because the pVAE baseline vastly relies on the latent optimisation for good performance, while the diffusion based methods do not.  We denote it as `no latent optimisation' setting when no/minimal latent optimisation is applied. In the ablation, we introduce the latent space optimisation to the diffusion models to explore its impact. Please be aware that latent space optimisation is different to the denoiser-decoder joint refinement.

\subsection{Baselines and Training}
\label{sec:exp:baselines}
The proposed framework is compared with representative pVAE-based and latent-diffusion-based inverse-design baselines. All methods use the same datasets, target conditions, surrogate evaluator, and Palabos verification protocol.

\paragraph{Structure-to-property surrogate.}
The surrogate $s_{\psi}$ is trained to predict the normalised/log-property vector directly from voxelised porous structures, as shown in Eq. (~\ref{eq:surrogate_prediction_method}). The surrogate is trained using Eq.~\eqref{eq:surrogate_loss}. It is used both as a fast evaluator during inverse design and as a differentiable feedback module during denoiser-decoder refinement. An independent voxel-space surrogate is used instead of the pVAE latent-property head to reduce evaluation bias. The surrogate is frozen during all experiments.

\paragraph{Baseline I: pVAE latent optimisation.}
The pVAE baseline follows property-aware VAE inverse design. Given a target condition $\mathbf{c}^{*}$, multiple latent codes are initialised from the pVAE prior and optimised using the frozen decoder and surrogate:
\begin{equation}
\mathbf{z}^{*}
=
\arg\min_{\mathbf{z}}
\left\|
s_{\psi}(D_{\phi}(\mathbf{z}))
-
\mathbf{c}^{*}
\right\|_2^2
+
\lambda_z\|\mathbf{z}\|_2^2.
\label{eq:pvae_baseline}
\end{equation}
The final structure is decoded as
\begin{equation}
\mathbf{x}_{\mathrm{pVAE}}^{*}
=
D_{\phi}(\mathbf{z}^{*}).
\label{eq:pvae_generated}
\end{equation}

\paragraph{Baseline II: latent diffusion with frozen decoder.}
The latent diffusion baseline is trained in the pVAE latent space using Eq.~\eqref{eq:diffusion_loss}. At inference time, the model samples target-conditioned latent candidates $\mathbf{z}_{\mathrm{diff}}$, which are decoded using the frozen pVAE decoder:
\begin{equation}
\mathbf{x}_{\mathrm{diff}}
=
D_{\phi}(\mathbf{z}_{\mathrm{diff}}).
\label{eq:diffusion_baseline_decode}
\end{equation}
When latent optimisaton is applied, as in Sec.~\ref{sec:ablation}, the latent representation is further optimised as
\begin{equation}
\mathbf{z}_{\mathrm{diff}}^{*}
=
\arg\min_{\mathbf{z}}
\left\|
s_{\psi}(D_{\phi}(\mathbf{z}))
-
\mathbf{c}^{*}
\right\|_2^2
+
\lambda_z
\left\|
\mathbf{z}
-
\mathbf{z}_{\mathrm{diff}}
\right\|_2^2.
\label{eq:diffusion_baseline_refinement}
\end{equation}
This baseline evaluates whether diffusion provides a stronger target-aware latent initialisation than random pVAE sampling while keeping the decoder fixed.

\paragraph{Our method.}
The proposed method starts from the trained conditional latent diffusion model and jointly refines the diffusion denoiser and pVAE decoder using Eq.~\eqref{eq:finetune_loss}. This refinement is designed to reduce the mismatch between diffusion-generated latent samples and the latent distribution originally observed by the pVAE decoder.

\subsection{LBM-Based Verification and Evaluation Metrics}
\label{sec:exp:metrics}

The surrogate provides differentiable property feedback during training and refinement, and the property is one option to be used for model evaluation. However, physical platforms can provide more physics-consistent evaluation. Therefore, the final evaluation is performed using Palabos-based LBM verification. For each generated sample, Palabos produces
\begin{equation}
\mathbf{y}_{\mathrm{LBM}}
=
[
n_F^{\mathrm{LBM}},
K_x^{\mathrm{LBM}},
K_y^{\mathrm{LBM}},
K_z^{\mathrm{LBM}}
]^T.
\label{eq:lbm_property}
\end{equation}

\paragraph{Palabos-verified relative error.}
The relative porosity error is
\begin{equation}
E_{n_F}
=
\frac{
\left|
n_F^{\mathrm{LBM}}-n_F^{*}
\right|
}
{
\left|n_F^{*}\right|+\delta
},
\label{eq:porosity_error}
\end{equation}
and the relative permeability error along direction $i\in\{x,y,z\}$ is
\begin{equation}
E_{K_i}
=
\frac{
\left|
K_i^{\mathrm{LBM}}-K_i^{*}
\right|
}
{
\left|K_i^{*}\right|+\delta
},
\label{eq:perm_error}
\end{equation}
where $\delta$ is a small constant for numerical stability.

The mean permeability error is
\begin{equation}
E_{K,\mathrm{mean}}
=
\frac{1}{3}
\left(
E_{K_x}
+
E_{K_y}
+
E_{K_z}
\right).
\label{eq:mean_perm_error}
\end{equation}

\paragraph{Multi-property evaluation score.}
The overall target-matching score is defined as
\begin{equation}
R
=
\sqrt{
E_{n_F}^2
+
E_{K_x}^2
+
E_{K_y}^2
+
E_{K_z}^2
}.
\label{eq:rank_score}
\end{equation}
Lower $R$ indicates closer agreement between the generated structure and the target properties.

\paragraph{Target controllability.}
To measure whether generated structures follow the requested target trends, we compute Pearson correlation between target and LBM-verified properties:
\begin{equation}
\rho_{n_F}
=
\mathrm{corr}
\left(
n_F^{*},
n_F^{\mathrm{LBM}}
\right),
\label{eq:corr_porosity}
\end{equation}
and
\begin{equation}
\rho_{K_i}
=
\mathrm{corr}
\left(
K_i^{*},
K_i^{\mathrm{LBM}}
\right),
\qquad
i\in\{x,y,z\}.
\label{eq:corr_perm}
\end{equation}
Higher correlation indicates stronger target-aware controllability.

\subsection{Results}
\label{sec:results}

All models are trained and evaluated on both the synthetic porous-media dataset and the real micro-CT dataset using an NVIDIA RTX 5090 GPU. For each evaluation setting, target property vectors are sampled and used to condition the generative models. Generated candidates are first evaluated using the frozen structure-to-property surrogate and are then verified using Palabos. We report both aggregate quantitative metrics and representative generated structures. Additional latent space optimisation results are provided in the ablation study.

\subsubsection{Results on Synthetic Data}
\label{sec:results:synthetic}

\providecommand{\porousimgw}{0.135\textwidth}
\providecommand{\porousvspacefig}{\\[-6pt]}

\begin{figure*}[htbp]
    \centering
    \includegraphics[width=\textwidth]{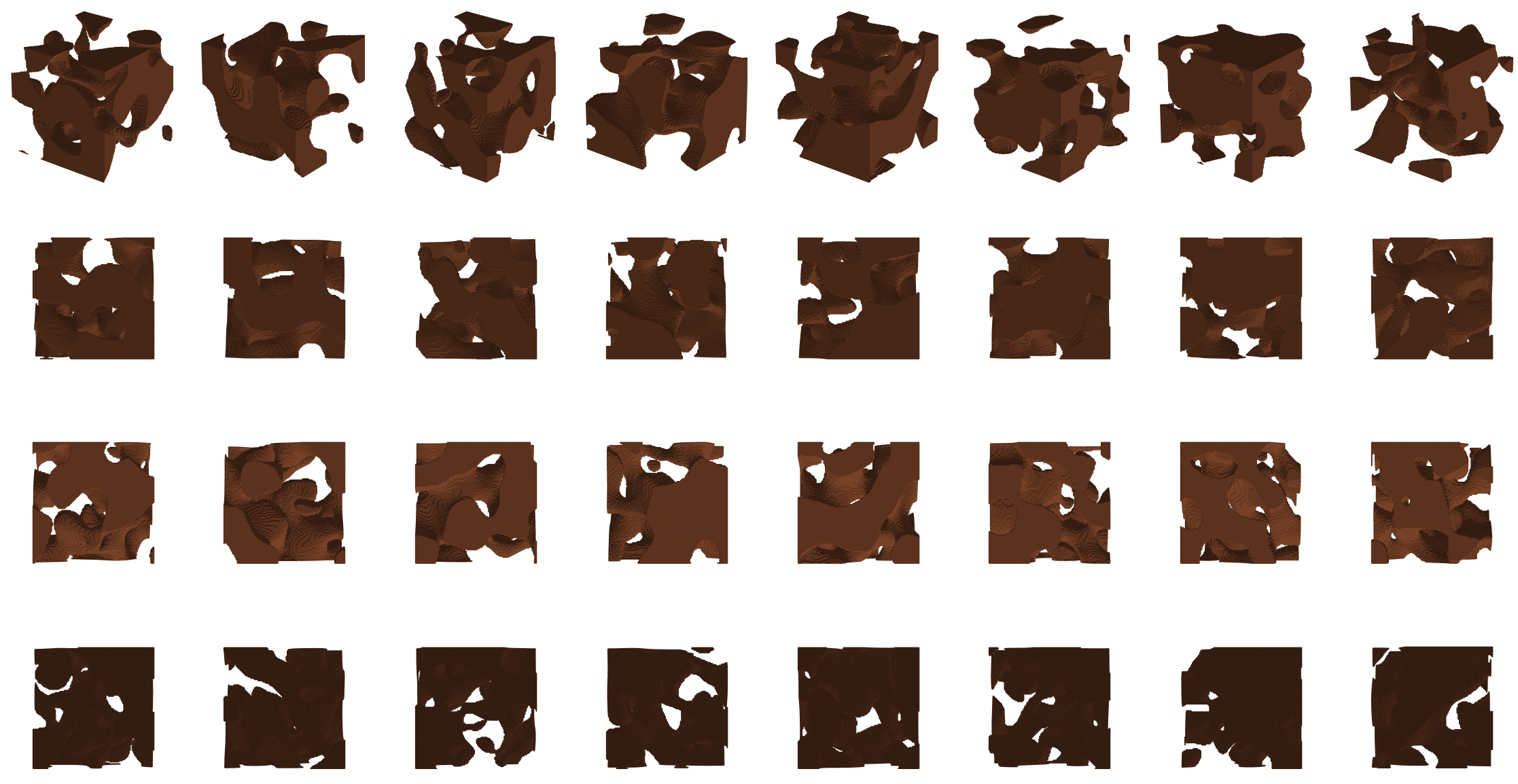}
    \caption{Porous structures generated by the pVAE method for randomly selected synthetic targets. Each column corresponds to one target, with isometric, front, side, and top views arranged from top to bottom.}
    \label{fig:sim_pvae_selected_8_targets}
\end{figure*}

\begin{figure*}[htbp]
    \centering
    \includegraphics[width=\textwidth]{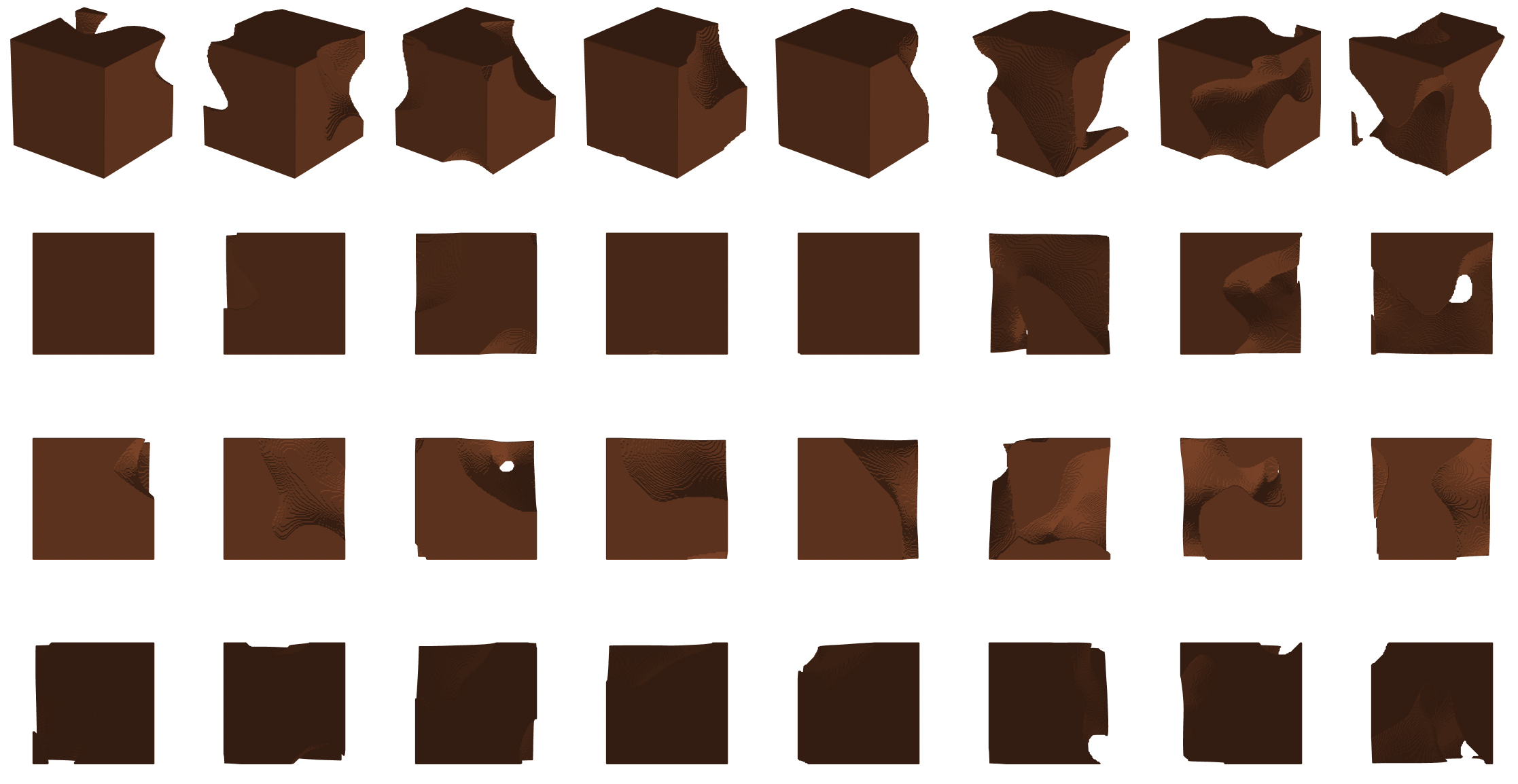}
    \caption{Porous structures generated by the latent diffusion method for randomly selected synthetic targets. Each column corresponds to one target, with isometric, front, side, and top views arranged from top to bottom.}
    \label{fig:sim_diffseed_selected_8_targets}
\end{figure*}

\begin{figure*}[htbp]
    \centering
    \includegraphics[width=\textwidth]{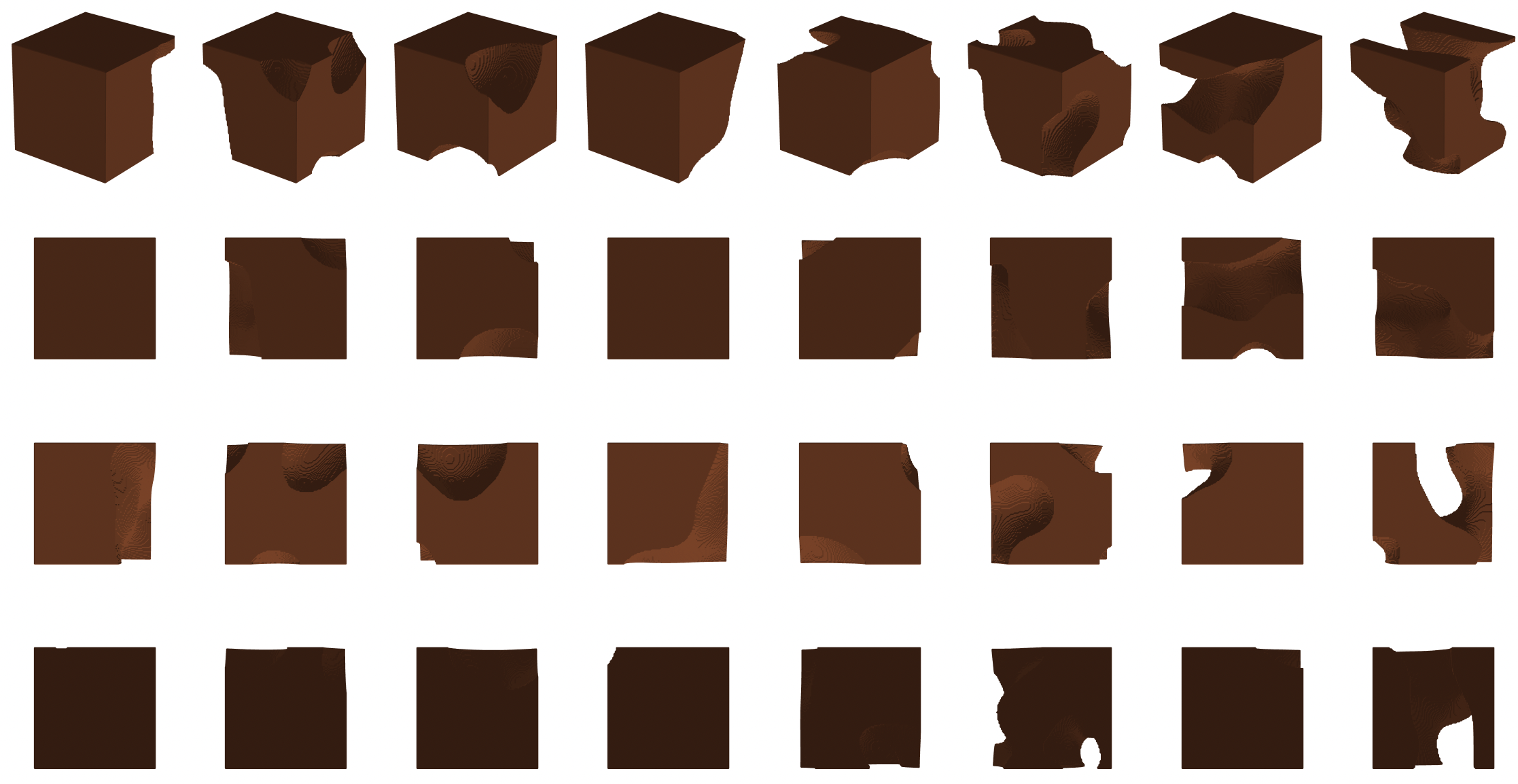}
    \caption{Porous structures generated by the proposed method for randomly selected synthetic targets. Each column corresponds to one target, with isometric, front, side, and top views arranged from top to bottom.}
    \label{fig:sim_joint_selected_8_targets}
\end{figure*}

Table~\ref{tab:sim3000_80_compare_step0} reports the Palabos-verified results on the synthetic dataset under the `no latent space optimisation' setting. In this setting, each method generates candidates directly from its learned sampling process, without additional latent-space optimisation after sampling. The pVAE baseline is evaluated with the minimum latent update required to produce a decoded candidate, while the latent diffusion baseline and the proposed method generate candidates directly from their conditional sampling processes.

The proposed method achieves the best mean score, median score, porosity error, mean permeability error, and most controllability correlations. The only exception is the Pearson correlation for \(K_x\), where the latent diffusion baseline and the proposed method perform similarly. Compared with the pVAE baseline, both diffusion-based methods substantially improve target controllability, indicating that conditional latent sampling provides a more effective mechanism for matching target physical properties than direct random latent search. The proposed method further improves over the latent diffusion baseline, suggesting that surrogate-guided denoiser-decoder refinement improves the compatibility between generated latent samples and physically meaningful decoded porous structures.

\begin{table*}[htbp]
\centering
\caption{LBM-verified comparison on the synthetic dataset under the `no latent space optimisation' setting. The pVAE baseline requires latent-space optimisation for inverse design and is evaluated with the minimum update step needed to produce a decoded candidate, while the diffusion-based methods generate candidates directly from conditional sampling. Lower error values are better, while higher correlation values indicate stronger target controllability. Green indicates the best result in each column.}
\label{tab:sim3000_80_compare_step0}
\small
\setlength{\tabcolsep}{3.0pt}
\renewcommand{\arraystretch}{1.08}
\resizebox{\textwidth}{!}{%
\begin{tabular}{llcccccccc}
\hline
Method & Steps &
Mean score $\downarrow$ &
Median score $\downarrow$ &
$n_F$ error $\downarrow$ &
Mean $K$ error $\downarrow$ &
Corr. $n_F$ $\uparrow$ &
Corr. $K_x$ $\uparrow$ &
Corr. $K_y$ $\uparrow$ &
Corr. $K_z$ $\uparrow$ \\
\hline
pVAE & 1 &
2.416 & 1.759 & 0.461 & 1.168 & 0.351 & 0.176 & 0.138 & 0.139 \\
Latent diffusion & - &
1.463 & 0.629 & 0.101 & 0.643 & 0.963 &
\cellcolor{green!25}0.961 & 0.941 & 0.940 \\
Ours & - &
\cellcolor{green!25}1.044 &
\cellcolor{green!25}0.533 &
\cellcolor{green!25}0.063 &
\cellcolor{green!25}0.474 &
\cellcolor{green!25}0.985 &
0.959 &
\cellcolor{green!25}0.966 &
\cellcolor{green!25}0.957 \\
\hline
\end{tabular}%
}
\end{table*}

Table~\ref{tab:simulation_selected_properties} provides representative target, surrogate-predicted, and Palabos-verified properties for selected generated structures. The pVAE baseline often produces structures with large deviations from the target porosity and permeability values. In contrast, latent diffusion produces substantially closer target matches, while the proposed method achieves the lowest score for all selected cases. This confirms that the proposed framework improves not only average performance but also per-target property consistency.

\begin{table*}[htbp]
\centering
\scriptsize
\setlength{\tabcolsep}{2.5pt}
\renewcommand{\arraystretch}{1.08}
\caption{Comparison between target properties, structure-to-property predicted properties, and Palabos-verified properties for selected generated porous structures from the synthetic dataset. \(n_F\) denotes pore fraction.}
\label{tab:simulation_selected_properties}
\resizebox{\textwidth}{!}{%
\begin{tabular}{cl|rrrr|rrrr|rrrr|r}
\toprule
\multirow{2}{*}{ID} & \multirow{2}{*}{Method} &
\multicolumn{4}{c|}{Target Properties} &
\multicolumn{4}{c|}{Structure-to-property Predictions} &
\multicolumn{4}{c|}{Palabos-verified Properties} &
\multirow{2}{*}{Score} \\
\cmidrule(lr){3-6}\cmidrule(lr){7-10}\cmidrule(lr){11-14}
& & $n_F$ & $K_x$ & $K_y$ & $K_z$
& $n_F$ & $K_x$ & $K_y$ & $K_z$
& $n_F$ & $K_x$ & $K_y$ & $K_z$ & \\
\midrule
\multirow{3}{*}{1}
& pVAE & 0.875 & 117.687 & 228.855 & 168.397 & 0.406 & 17.175 & 10.298 & 7.566 & 0.281 & 2.673 & 0.796 & 0.542 & 1.845 \\
& Latent diffusion & 0.875 & 117.687 & 228.855 & 168.397 & 0.849 & 100.813 & 205.168 & 183.697 & 0.816 & 111.421 & 175.623 & 187.364 & 0.272 \\
& Ours & 0.875 & 117.687 & 228.855 & 168.397 & 0.855 & 100.302 & 188.197 & 139.339 & 0.862 & 111.028 & 234.246 & 165.783 & \textbf{0.065} \\
\midrule
\multirow{3}{*}{2}
& pVAE & 0.878 & 224.861 & 77.083 & 239.821 & 0.375 & 11.418 & 9.875 & 26.398 & 0.326 & 3.615 & 2.411 & 2.754 & 1.811 \\
& Latent diffusion & 0.878 & 224.861 & 77.083 & 239.821 & 0.859 & 187.219 & 84.603 & 235.333 & 0.866 & 245.116 & 91.869 & 250.609 & 0.217 \\
& Ours & 0.878 & 224.861 & 77.083 & 239.821 & 0.862 & 206.237 & 89.079 & 227.428 & 0.851 & 241.310 & 78.662 & 232.455 & \textbf{0.088} \\
\midrule
\multirow{3}{*}{3}
& pVAE & 0.910 & 256.143 & 207.859 & 222.744 & 0.368 & 15.056 & 7.723 & 11.848 & 0.280 & 3.264 & 0.561 & 2.745 & 1.851 \\
& Latent diffusion & 0.910 & 256.143 & 207.859 & 222.744 & 0.867 & 214.016 & 195.375 & 213.649 & 0.859 & 192.145 & 221.089 & 185.535 & 0.312 \\
& Ours & 0.910 & 256.143 & 207.859 & 222.744 & 0.904 & 220.319 & 212.285 & 225.264 & 0.895 & 246.565 & 209.390 & 201.243 & \textbf{0.105} \\
\midrule
\multirow{3}{*}{4}
& pVAE & 0.934 & 259.180 & 249.850 & 297.041 & 0.386 & 16.601 & 15.597 & 14.409 & 0.334 & 3.640 & 3.026 & 3.984 & 1.826 \\
& Latent diffusion & 0.934 & 259.180 & 249.850 & 297.041 & 0.943 & 265.442 & 237.413 & 282.057 & 0.936 & 317.784 & 202.094 & 259.254 & 0.322 \\
& Ours & 0.934 & 259.180 & 249.850 & 297.041 & 0.939 & 249.215 & 259.628 & 278.530 & 0.941 & 289.683 & 278.780 & 284.199 & \textbf{0.171} \\
\midrule
\multirow{3}{*}{5}
& pVAE & 0.848 & 159.312 & 115.907 & 193.426 & 0.420 & 13.440 & 6.031 & 18.698 & 0.326 & 3.944 & 1.724 & 1.991 & 1.811 \\
& Latent diffusion & 0.848 & 159.312 & 115.907 & 193.426 & 0.881 & 171.213 & 128.412 & 209.799 & 0.884 & 202.090 & 170.276 & 236.629 & 0.586 \\
& Ours & 0.848 & 159.312 & 115.907 & 193.426 & 0.855 & 181.886 & 115.072 & 186.911 & 0.857 & 187.127 & 100.546 & 197.972 & \textbf{0.221} \\
\midrule
\multirow{3}{*}{6}
& pVAE & 0.613 & 26.383 & 89.462 & 94.756 & 0.381 & 7.039 & 31.923 & 10.495 & 0.304 & 1.471 & 2.148 & 3.138 & 1.741 \\
& Latent diffusion & 0.613 & 26.383 & 89.462 & 94.756 & 0.593 & 19.078 & 93.047 & 54.702 & 0.604 & 45.665 & 73.350 & 62.330 & 0.827 \\
& Ours & 0.613 & 26.383 & 89.462 & 94.756 & 0.643 & 19.047 & 105.179 & 109.422 & 0.683 & 22.476 & 92.771 & 79.589 & \textbf{0.249} \\
\midrule
\multirow{3}{*}{7}
& pVAE & 0.804 & 74.783 & 132.944 & 119.116 & 0.384 & 4.432 & 12.551 & 19.979 & 0.296 & 2.026 & 1.336 & 1.495 & 1.817 \\
& Latent diffusion & 0.804 & 74.783 & 132.944 & 119.116 & 0.754 & 94.037 & 105.831 & 107.379 & 0.760 & 82.806 & 85.905 & 119.638 & 0.374 \\
& Ours & 0.804 & 74.783 & 132.944 & 119.116 & 0.833 & 73.049 & 181.073 & 158.302 & 0.799 & 89.401 & 145.305 & 103.401 & \textbf{0.254} \\
\midrule
\multirow{3}{*}{8}
& pVAE & 0.447 & 19.021 & 26.187 & 46.286 & 0.412 & 11.306 & 21.206 & 13.006 & 0.300 & 1.825 & 3.184 & 0.666 & 1.633 \\
& Latent diffusion & 0.447 & 19.021 & 26.187 & 46.286 & 0.547 & 21.827 & 26.291 & 36.406 & 0.532 & 4.307 & 32.363 & 28.209 & 0.918 \\
& Ours & 0.447 & 19.021 & 26.187 & 46.286 & 0.498 & 20.659 & 30.740 & 33.725 & 0.468 & 19.006 & 28.994 & 35.817 & \textbf{0.255} \\
\bottomrule
\end{tabular}
}
\end{table*}

Representative generated samples are shown in Figs.~\ref{fig:sim_pvae_selected_8_targets},~\ref{fig:sim_diffseed_selected_8_targets}, and~\ref{fig:sim_joint_selected_8_targets}. Each column corresponds to one target, with isometric, front, side, and top views arranged from top to bottom. The pVAE baseline generate samples that do not consistently follow the requested target-property trends. The latent diffusion baseline improves target alignment, while the proposed method further improves simulator-verified property matching while maintaining realistic porous geometry.

\subsubsection{Results on Real Micro-CT Data}
\label{sec:results:real}

Table~\ref{tab:central80_compare_step0} reports the Palabos-verified results on the real micro-CT dataset. The proposed method achieves the best mean score, median score, mean permeability error, and all Pearson correlation metrics. This indicates that the generated structures not only reduce average property mismatch but also follow the requested target-property trends more consistently.

The pVAE baseline achieves the lowest porosity error, but its correlations are weak, indicating limited controllability across different target conditions. The latent diffusion baseline improves controllability relative to pVAE, but its overall score and permeability errors remain higher than those of the proposed method. These results suggest that conditioning alone is insufficient when the decoder remains fixed. By jointly refining the denoiser and decoder using differentiable surrogate feedback, the proposed method better aligns diffusion-generated latent samples with physically meaningful decoded porous structures.

The higher porosity error of the proposed method reflects a trade-off between global pore fraction and directional permeability consistency. Permeability is more sensitive to pore connectivity, channel continuity, and anisotropic flow pathways, whereas porosity is primarily a global volumetric statistic. During joint refinement, the model may therefore introduce small changes in pore volume fraction to improve transport-relevant structures and better match the target permeability.

\begin{table*}[htbp]
\centering
\caption{LBM-verified comparison on the real micro-CT dataset under the `no latent space optimisation' setting. The pVAE baseline requires latent-space optimisation for inverse design and is evaluated with the minimum update step needed to produce a decoded candidate, while the diffusion-based methods generate candidates directly from conditional sampling. Lower error values are better, while higher correlation values indicate stronger target controllability. Green indicates the best result in each column.}
\label{tab:central80_compare_step0}
\small
\setlength{\tabcolsep}{3.0pt}
\renewcommand{\arraystretch}{1.08}
\resizebox{\textwidth}{!}{%
\begin{tabular}{llcccccccc}
\hline
Method & Steps &
Mean score $\downarrow$ &
Median score $\downarrow$ &
$n_F$ error $\downarrow$ &
Mean $K$ error $\downarrow$ &
Corr. $n_F$ $\uparrow$ &
Corr. $K_x$ $\uparrow$ &
Corr. $K_y$ $\uparrow$ &
Corr. $K_z$ $\uparrow$ \\
\hline
pVAE & 1 &
0.309 &
0.314 &
\cellcolor{green!25}0.019 &
0.157 &
-0.056 &
0.048 &
-0.017 &
0.171 \\
Latent diffusion & - &
0.554 &
0.515 &
0.024 &
0.273 &
0.717 &
0.391 &
0.240 &
0.282 \\
Ours & - &
\cellcolor{green!25}0.306 &
\cellcolor{green!25}0.285 &
0.106 &
\cellcolor{green!25}0.147 &
\cellcolor{green!25}0.821 &
\cellcolor{green!25}0.758 &
\cellcolor{green!25}0.727 &
\cellcolor{green!25}0.740 \\
\hline
\end{tabular}%
}
\end{table*}

Table~\ref{tab:physics_selected_properties} shows representative target, surrogate-predicted, and Palabos-verified properties for selected real micro-CT cases. Although the pVAE and latent diffusion baselines can generate visually plausible porous structures, their Palabos-verified properties often deviate from the desired targets, especially for directional permeability. This highlights an important challenge in porous-media inverse design: visual realism alone does not guarantee physical correctness. In contrast, the proposed method achieves lower target-matching scores for the selected cases, indicating stronger physical consistency.

\begin{table*}[htbp]
\centering
\scriptsize
\setlength{\tabcolsep}{2.5pt}
\renewcommand{\arraystretch}{1.08}
\caption{Comparison between target properties, structure-to-property predicted properties, and Palabos-verified properties for selected generated porous structures from the real micro-CT dataset. \(n_F\) denotes pore fraction.}
\label{tab:physics_selected_properties}
\resizebox{\textwidth}{!}{%
\begin{tabular}{cl|rrrr|rrrr|rrrr|r}
\toprule
\multirow{2}{*}{ID} & \multirow{2}{*}{Method} &
\multicolumn{4}{c|}{Target Properties} &
\multicolumn{4}{c|}{Structure-to-property Predictions} &
\multicolumn{4}{c|}{Palabos-verified Properties} &
\multirow{2}{*}{Score} \\
\cmidrule(lr){3-6}\cmidrule(lr){7-10}\cmidrule(lr){11-14}
& & $n_F$ & $K_x$ & $K_y$ & $K_z$
& $n_F$ & $K_x$ & $K_y$ & $K_z$
& $n_F$ & $K_x$ & $K_y$ & $K_z$ & \\
\midrule
\multirow{3}{*}{1}
& pVAE & 0.737 & 13.591 & 11.633 & 9.599 & 0.821 & 21.768 & 17.072 & 18.316 & 0.713 & 9.946 & 9.194 & 9.113 & 0.346 \\
& Latent diffusion & 0.737 & 13.591 & 11.633 & 9.599 & 0.842 & 24.210 & 16.801 & 18.243 & 0.755 & 21.662 & 12.412 & 14.546 & 0.789 \\
& Ours & 0.737 & 13.591 & 11.633 & 9.599 & 0.740 & 13.698 & 12.208 & 9.537 & 0.659 & 14.251 & 10.853 & 9.750 & \textbf{0.135} \\
\midrule
\multirow{3}{*}{2}
& pVAE & 0.688 & 12.458 & 10.152 & 7.167 & 0.824 & 19.790 & 16.968 & 18.718 & 0.721 & 10.042 & 8.646 & 9.159 & 0.373 \\
& Latent diffusion & 0.688 & 12.458 & 10.152 & 7.167 & 0.805 & 18.712 & 16.234 & 14.986 & 0.674 & 11.479 & 5.996 & 9.886 & 0.564 \\
& Ours & 0.688 & 12.458 & 10.152 & 7.167 & 0.701 & 12.140 & 10.228 & 7.282 & 0.619 & 12.103 & 10.699 & 7.858 & \textbf{0.152} \\
\midrule
\multirow{3}{*}{3}
& pVAE & 0.738 & 13.553 & 10.751 & 11.323 & 0.825 & 20.544 & 17.832 & 19.087 & 0.717 & 9.553 & 8.491 & 9.284 & 0.406 \\
& Latent diffusion & 0.738 & 13.553 & 10.751 & 11.323 & 0.837 & 25.126 & 18.022 & 18.903 & 0.736 & 16.092 & 12.259 & 11.522 & 0.235 \\
& Ours & 0.738 & 13.553 & 10.751 & 11.323 & 0.749 & 13.922 & 10.942 & 11.559 & 0.668 & 13.405 & 10.418 & 12.852 & \textbf{0.168} \\
\midrule
\multirow{3}{*}{4}
& pVAE & 0.714 & 12.807 & 9.295 & 11.094 & 0.825 & 20.712 & 15.674 & 19.761 & 0.721 & 9.560 & 10.177 & 12.167 & 0.288 \\
& Latent diffusion & 0.714 & 12.807 & 9.295 & 11.094 & 0.812 & 19.800 & 14.212 & 18.158 & 0.697 & 13.099 & 10.529 & 14.367 & 0.325 \\
& Ours & 0.714 & 12.807 & 9.295 & 11.094 & 0.719 & 12.469 & 9.862 & 10.682 & 0.637 & 11.949 & 10.262 & 10.356 & \textbf{0.177} \\
\midrule
\multirow{3}{*}{5}
& pVAE & 0.724 & 11.476 & 7.443 & 10.862 & 0.824 & 21.735 & 17.198 & 18.371 & 0.719 & 10.943 & 9.142 & 10.494 & 0.235 \\
& Latent diffusion & 0.724 & 11.476 & 7.443 & 10.862 & 0.821 & 18.824 & 16.678 & 18.029 & 0.705 & 9.429 & 8.269 & 11.971 & 0.235 \\
& Ours & 0.724 & 11.476 & 7.443 & 10.862 & 0.715 & 11.079 & 7.691 & 10.724 & 0.633 & 12.460 & 6.722 & 11.146 & \textbf{0.182} \\
\midrule
\multirow{3}{*}{6}
& pVAE & 0.734 & 10.934 & 7.248 & 10.715 & 0.821 & 17.898 & 17.956 & 16.134 & 0.709 & 7.058 & 8.762 & 7.382 & 0.517 \\
& Latent diffusion & 0.734 & 10.934 & 7.248 & 10.715 & 0.816 & 22.235 & 14.322 & 19.001 & 0.688 & 19.561 & 8.240 & 7.734 & 0.850 \\
& Ours & 0.734 & 10.934 & 7.248 & 10.715 & 0.732 & 10.569 & 7.170 & 11.023 & 0.657 & 9.395 & 6.849 & 10.550 & \textbf{0.185} \\
\midrule
\multirow{3}{*}{7}
& pVAE & 0.724 & 11.078 & 10.230 & 11.657 & 0.821 & 19.946 & 17.326 & 19.005 & 0.701 & 8.569 & 9.243 & 8.851 & 0.346 \\
& Latent diffusion & 0.724 & 11.078 & 10.230 & 11.657 & 0.831 & 19.159 & 18.069 & 19.548 & 0.723 & 7.974 & 13.246 & 12.900 & 0.420 \\
& Ours & 0.724 & 11.078 & 10.230 & 11.657 & 0.725 & 10.690 & 9.714 & 11.116 & 0.653 & 10.329 & 11.475 & 10.815 & \textbf{0.185} \\
\midrule
\multirow{3}{*}{8}
& pVAE & 0.690 & 10.906 & 10.131 & 8.355 & 0.823 & 20.781 & 17.922 & 20.679 & 0.723 & 10.268 & 8.890 & 9.778 & 0.223 \\
& Latent diffusion & 0.690 & 10.906 & 10.131 & 8.355 & 0.803 & 18.842 & 13.077 & 13.126 & 0.676 & 10.079 & 10.894 & 5.534 & 0.355 \\
& Ours & 0.690 & 10.906 & 10.131 & 8.355 & 0.703 & 11.099 & 10.165 & 8.782 & 0.622 & 11.259 & 8.714 & 8.992 & \textbf{0.190} \\
\bottomrule
\end{tabular}
}
\end{table*}

\begin{figure*}[htbp]
    \centering
    \includegraphics[width=\textwidth]{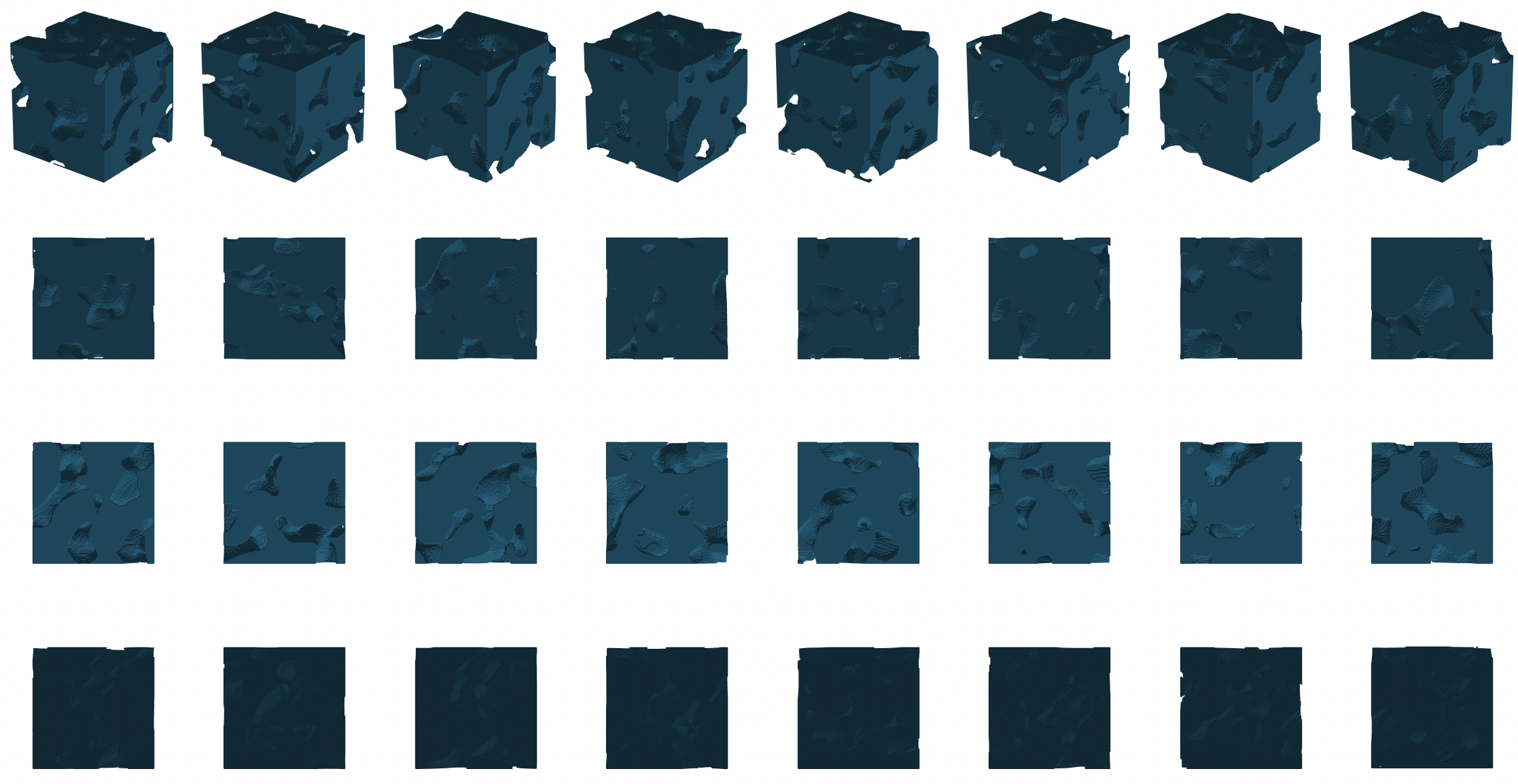}
    \caption{Porous structures generated by the pVAE method for randomly selected real micro-CT targets. Each column corresponds to one target, with isometric, front, side, and top views arranged from top to bottom.}
    \label{fig:pvae_selected_8_targets}
\end{figure*}

\begin{figure*}[htbp]
    \centering
    \includegraphics[width=\textwidth]{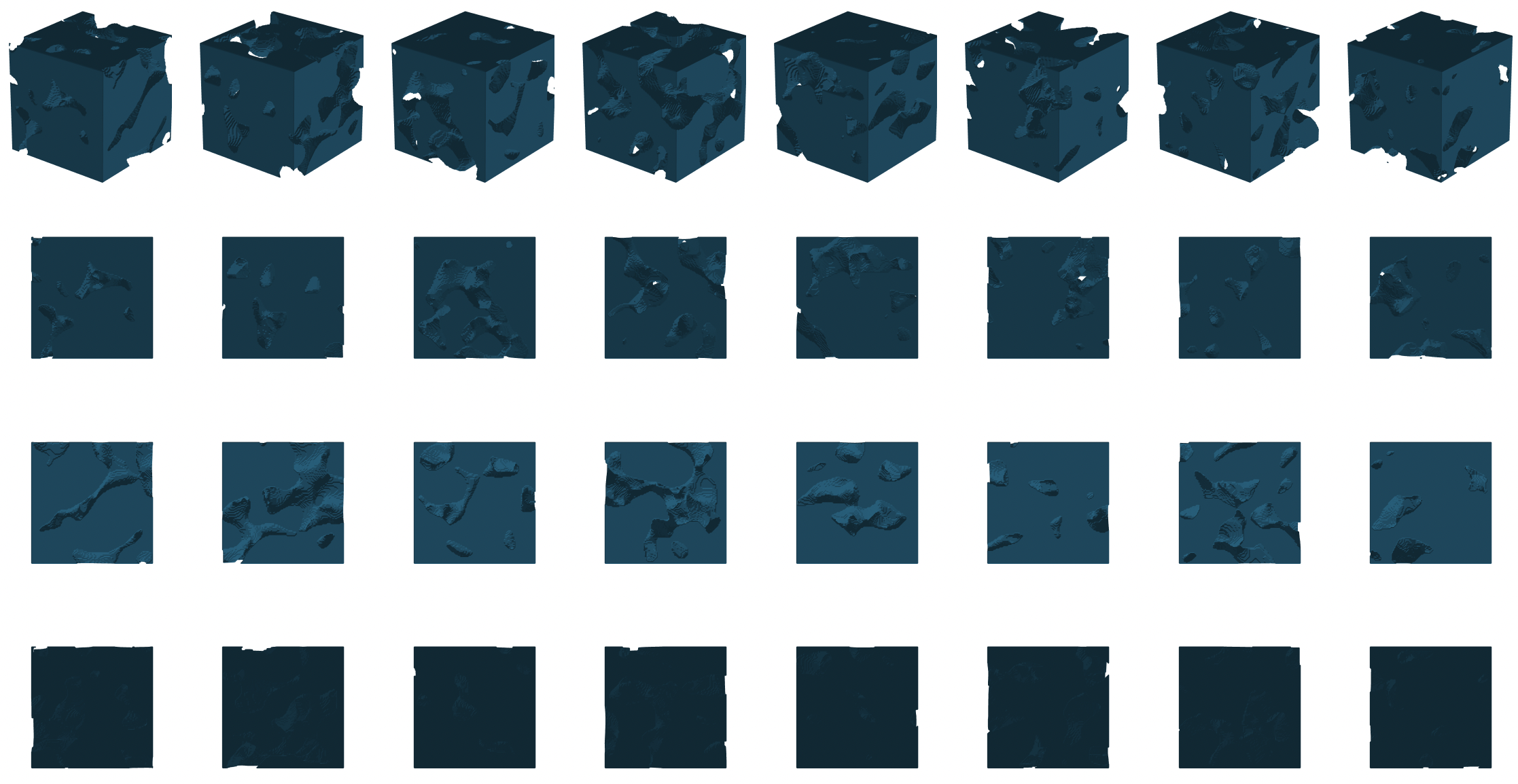}
    \caption{Porous structures generated by the latent diffusion method for randomly selected real micro-CT targets. Each column corresponds to one target, with isometric, front, side, and top views arranged from top to bottom.}
    \label{fig:diffseed_selected_8_targets}
\end{figure*}

\begin{figure*}[htbp]
    \centering
    \includegraphics[width=\textwidth]{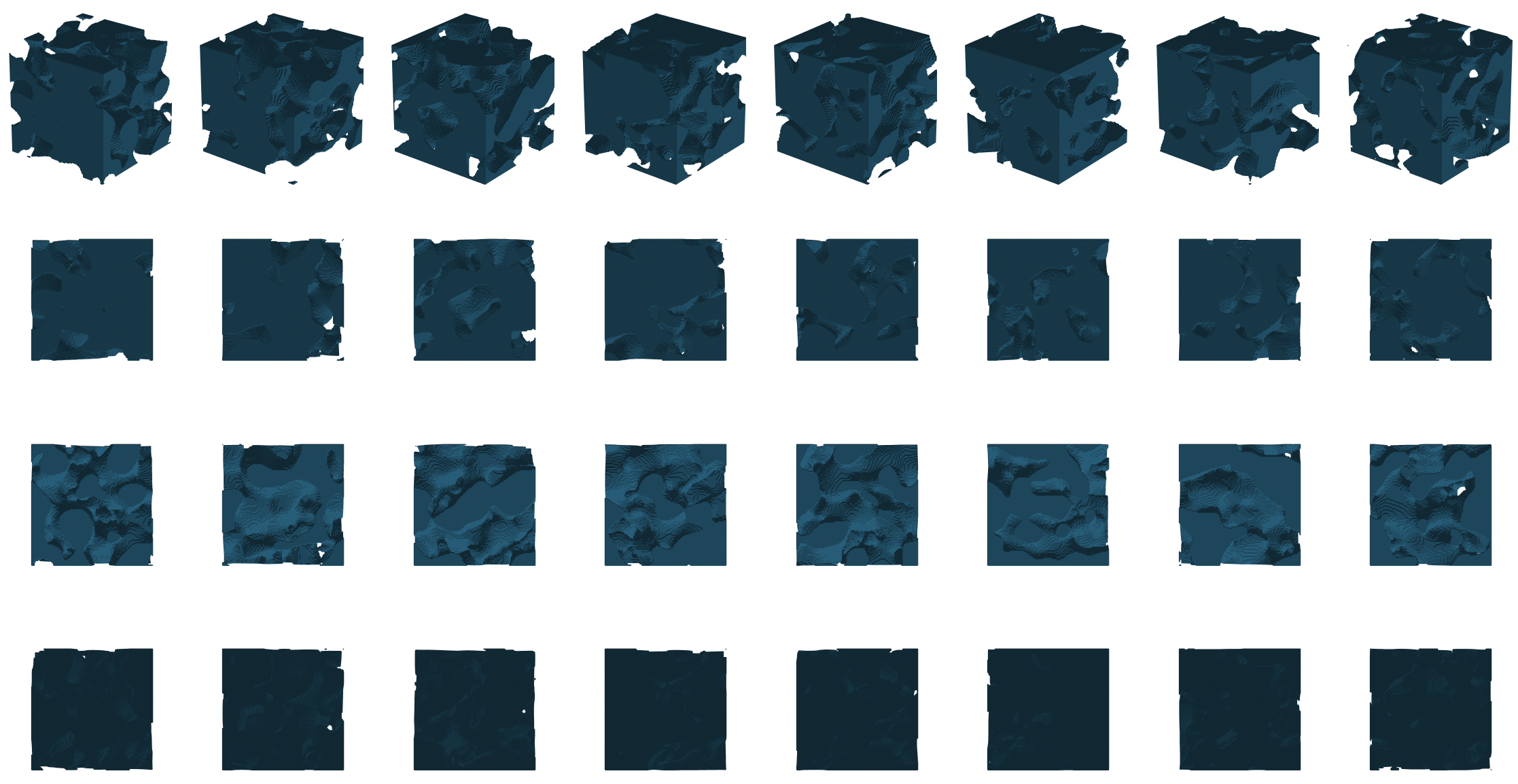}
    \caption{Porous structures generated by the proposed method for randomly selected real micro-CT targets. Each column corresponds to one target, with isometric, front, side, and top views arranged from top to bottom.}
    \label{fig:joint_selected_8_targets}
\end{figure*}

Representative structures are shown in Figs.~\ref{fig:pvae_selected_8_targets},~\ref{fig:diffseed_selected_8_targets}, and~\ref{fig:joint_selected_8_targets}. The proposed method preserves realistic porous geometry while producing structures whose verified transport behaviour is more consistent with the prescribed targets.

\subsection{Ablation}
\label{sec:ablation}

\subsubsection{Effect of Latent Space Optimisation}
\label{sec:abl:finetune}

Tables~\ref{tab:sim3000_80_compare_refinement} and~\ref{tab:central80_compare_refinement} evaluate the effect of latent-space optimisation steps on the synthetic and real micro-CT datasets, respectively. Following the optimisation strategy used in the pVAE inverse-design framework~\cite{nguyen2026deep}, we perform additional latent space optimisation for 10, 30, and 50 steps before decoding and Palabos verification. This experiment is designed to evaluate whether latent optimisation can improve target matching for the latent diffusion baseline and the proposed joint denoiser-decoder framework.

\begin{table*}[hptb]
\centering
\caption{LBM-verified comparison on the synthetic dataset with latent-space optimisation. We follow the latent optimisation paradigm used by the pVAE-based inverse-design baseline and evaluate 10, 30, and 50 optimisation steps. Lower error values are better, while higher correlation values indicate stronger target controllability. Green, blue, and red indicate the best, second-best, and third-best results in each column, respectively.}
\label{tab:sim3000_80_compare_refinement}
\small
\setlength{\tabcolsep}{3.0pt}
\renewcommand{\arraystretch}{1.08}
\resizebox{\textwidth}{!}{%
\begin{tabular}{llcccccccc}
\hline
Method & Steps &
Mean score $\downarrow$ &
Median score $\downarrow$ &
$n_F$ error $\downarrow$ &
Mean $K$ error $\downarrow$ &
Corr. $n_F$ $\uparrow$ &
Corr. $K_x$ $\uparrow$ &
Corr. $K_y$ $\uparrow$ &
Corr. $K_z$ $\uparrow$ \\
\hline
pVAE & 10 &
1.848 & 1.470 & 0.231 & 0.947 & 0.896 & 0.698 & 0.697 & 0.658 \\
Latent diffusion & 10 &
\cellcolor{red!20}1.217 &
\cellcolor{blue!20}0.564 &
0.069 &
\cellcolor{red!20}0.569 &
0.978 &
\cellcolor{red!20}0.961 &
\cellcolor{blue!20}0.953 &
0.911 \\
Ours & 10 &
\cellcolor{blue!20}1.046 &
\cellcolor{green!25}0.541 &
0.048 &
\cellcolor{blue!20}0.495 &
0.992 &
\cellcolor{green!25}0.975 &
\cellcolor{green!25}0.956 &
\cellcolor{green!25}0.968 \\
\hline
pVAE & 30 &
2.756 & 1.313 & 0.136 & 1.262 & 0.979 & 0.897 & 0.811 & 0.831 \\
Latent diffusion & 30 &
1.249 & 0.651 &
\cellcolor{red!20}0.044 &
0.584 &
\cellcolor{green!25}0.993 &
0.949 &
0.935 &
0.905 \\
Ours & 30 &
1.821 & 0.605 &
\cellcolor{blue!20}0.042 &
0.771 &
\cellcolor{blue!20}0.993 &
\cellcolor{blue!20}0.962 &
0.930 &
0.920 \\
\hline
pVAE & 50 &
2.376 & 1.288 & 0.108 & 1.151 & 0.984 & 0.928 & 0.881 & \cellcolor{blue!20}0.932 \\
Latent diffusion & 50 &
1.629 & 0.695 &
\cellcolor{green!25}0.040 &
0.726 &
\cellcolor{red!20}0.993 &
0.938 &
0.897 &
0.908 \\
Ours & 50 &
\cellcolor{green!25}0.980 &
\cellcolor{red!20}0.572 &
0.052 &
\cellcolor{green!25}0.464 &
0.989 &
0.928 &
\cellcolor{red!20}0.929 &
\cellcolor{red!20}0.923 \\
\hline
\end{tabular}%
}
\end{table*}

\begin{table*}[hptb]
\centering
\caption{LBM-verified comparison on the real micro-CT dataset with latent-space optimisation. We follow the latent optimisation paradigm used by the pVAE-based inverse-design baseline and evaluate 10, 30, and 50 optimisation steps. Lower error values are better, while higher correlation values indicate stronger target controllability. Green, blue, and red indicate the best, second-best, and third-best results in each column, respectively.}
\label{tab:central80_compare_refinement}
\small
\setlength{\tabcolsep}{3.0pt}
\renewcommand{\arraystretch}{1.08}
\resizebox{\textwidth}{!}{%
\begin{tabular}{llcccccccc}
\hline
Method & Steps &
Mean score $\downarrow$ &
Median score $\downarrow$ &
$n_F$ error $\downarrow$ &
Mean $K$ error $\downarrow$ &
Corr. $n_F$ $\uparrow$ &
Corr. $K_x$ $\uparrow$ &
Corr. $K_y$ $\uparrow$ &
Corr. $K_z$ $\uparrow$ \\
\hline
pVAE & 10 &
0.555 &
0.546 &
0.116 &
0.290 &
0.688 &
0.462 &
0.382 &
0.303 \\
Latent diffusion & 10 &
0.364 &
0.343 &
\cellcolor{green!25}0.076 &
0.174 &
0.710 &
0.514 &
0.374 &
0.461 \\
Ours & 10 &
\cellcolor{blue!20}0.323 &
0.312 &
0.107 &
\cellcolor{blue!20}0.153 &
\cellcolor{red!20}0.915 &
0.634 &
0.606 &
\cellcolor{red!20}0.738 \\
\hline
pVAE & 30 &
0.473 &
0.492 &
0.145 &
0.237 &
0.912 &
\cellcolor{red!20}0.644 &
\cellcolor{green!25}0.746 &
0.671 \\
Diffusion seed & 30 &
0.418 &
0.406 &
0.124 &
0.196 &
0.778 &
0.525 &
0.527 &
0.590 \\
Joint decoder & 30 &
\cellcolor{green!25}0.309 &
\cellcolor{blue!20}0.300 &
0.105 &
\cellcolor{green!25}0.144 &
\cellcolor{blue!20}0.926 &
\cellcolor{blue!20}0.704 &
\cellcolor{blue!20}0.664 &
\cellcolor{blue!20}0.775 \\
\hline
pVAE & 50 &
0.425 &
0.407 &
0.146 &
0.208 &
0.905 &
0.710 &
0.621 &
0.727 \\
Diffusion seed & 50 &
0.429 &
0.389 &
0.137 &
0.200 &
0.903 &
0.625 &
\cellcolor{red!20}0.630 &
0.650 \\
Joint decoder & 50 &
\cellcolor{red!20}0.313 &
\cellcolor{green!25}0.297 &
\cellcolor{blue!20}0.104 &
\cellcolor{red!20}0.146 &
\cellcolor{green!25}0.932 &
\cellcolor{green!25}0.785 &
\cellcolor{green!25}0.710 &
\cellcolor{green!25}0.818 \\
\hline
\end{tabular}%
}
\end{table*}

Overall, latent optimisation improves controllability for all methods, particularly on the real micro-CT dataset. However, the magnitude of improvement differs substantially across methods. On the synthetic dataset, the pVAE baseline benefits from additional optimisation steps, especially in porosity and permeability correlations, but its overall mean score and mean permeability error remain substantially worse than the diffusion-based approaches. This indicates that direct latent optimisation can partially improve target matching, but the pVAE latent space alone is insufficient for reliable high-dimensional inverse design.

The latent diffusion baseline achieves stronger performance than pVAE across most metrics, confirming that diffusion-generated latent initialisation provides a better starting point for inverse design than random pVAE latent sampling. However, increasing optimisation steps does not always improve the final score consistently. This suggests that optimisation within a frozen decoder space may gradually move latent samples away from the decoder's most physically reliable region.

The proposed method achieves the best overall balance between target accuracy and controllability. On the synthetic dataset, it achieves the best overall mean score at 50 optimisation steps and produces strong or near-strong Pearson correlations across all physical properties. On the real micro-CT dataset, it consistently achieves the best or near-best mean score, median score, mean permeability error, and permeability correlations across optimisation settings. These results suggest that joint denoiser-decoder refinement is especially beneficial when dealing with real porous structures, where the latent distribution is more complex and decoder mismatch becomes more pronounced.

An additional observation is that the porosity error of the proposed method is occasionally slightly higher than that of the latent diffusion baseline. This is likely caused by the stronger optimisation emphasis on directional permeability consistency. Permeability depends heavily on pore connectivity, and anisotropic flow pathways, whereas porosity is primarily a global volumetric statistic. During joint refinement, the model may introduce small changes in pore volume fraction to better preserve transport-relevant structures required for matching directional permeability. Despite this trade-off, the proposed method maintains competitive porosity accuracy while achieving stronger overall physical-property controllability and simulator-verified consistency.

\subsubsection{Effect of Latent Space Resolution}
\label{sec:abl:latent}

All three methods rely on a pVAE to construct a compact latent representation for downstream porous-structure generation and inverse design. The quality of this latent representation is therefore critical, as it directly affects reconstruction fidelity, latent controllability, and the effectiveness of subsequent diffusion-based generation.

\begin{figure}[tbph]
    \centering
    \includegraphics[width=1.0\linewidth]{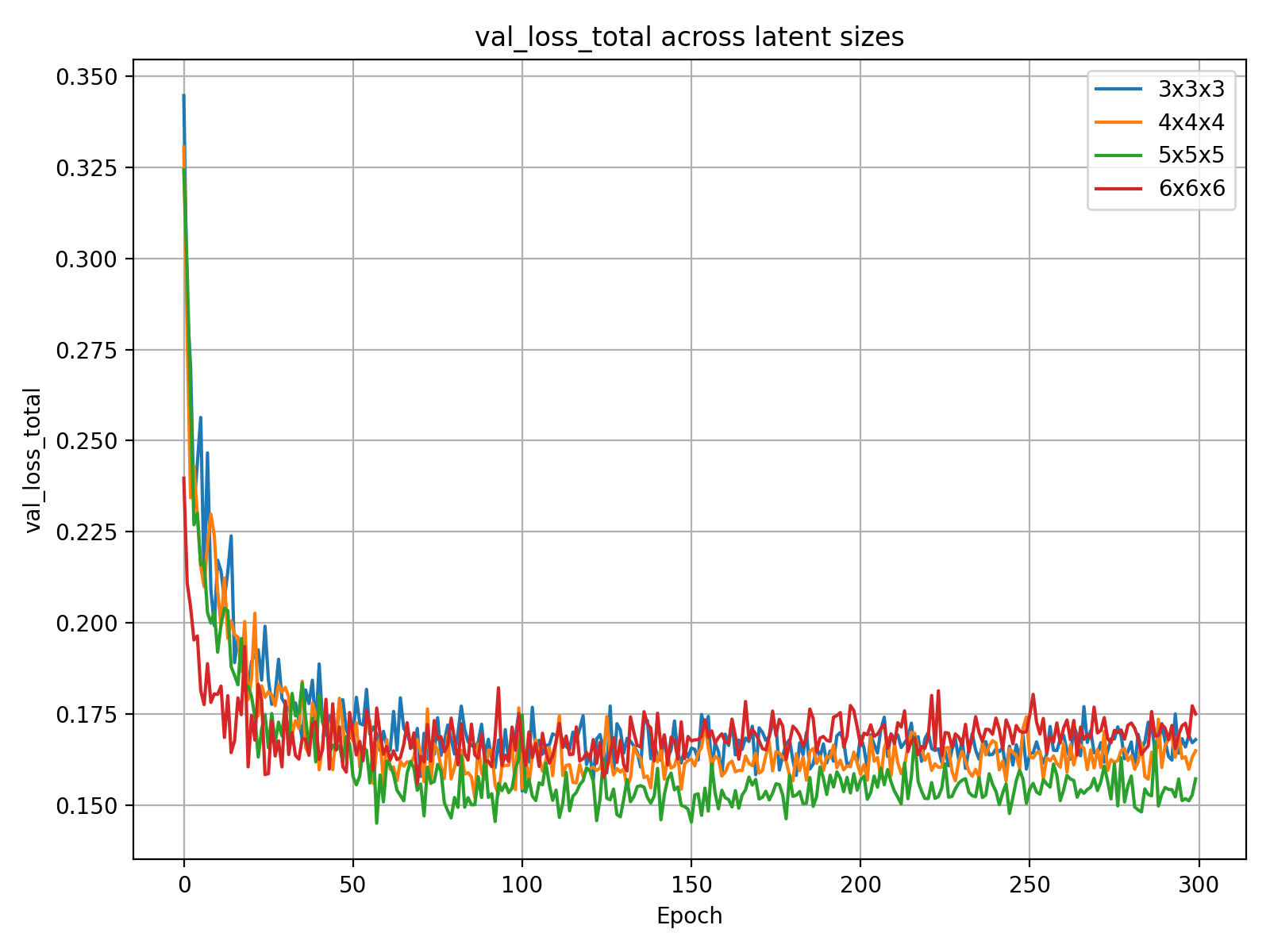}
    \caption{Comparison of the impact of latent space dimension on property prediction and decoder performance.}
    \label{fig:latentresolution}
\end{figure}

To determine an appropriate latent representation size, we evaluate latent tensor resolutions of \(3^3\), \(4^3\), \(5^3\), and \(6^3\) using the real micro-CT dataset. The trained encoder maps porous structures into latent tensors of different spatial resolutions, while the decoder reconstructs the voxelised geometry from the latent representation. Figure~\ref{fig:latentresolution} shows that a latent resolution of \(5^3\) achieves the best overall performance. Smaller latent spaces such as \(3^3\) and \(4^3\) introduce excessive information compression and reduce reconstruction fidelity, whereas larger latent spaces such as \(6^3\) increase latent complexity without providing consistent performance improvements. Based on these observations, a latent tensor resolution of \(5^3\) is adopted for all subsequent experiments. The resulting trained pVAE decoder is then shared across all compared methods to map latent representations back into voxelised porous structures.

Overall, these ablation results demonstrate that latent optimisation alone is insufficient for robust inverse design unless the latent distribution, denoiser, and decoder remain physically compatible. The proposed joint denoiser-decoder refinement provides a more stable and controllable optimisation space, enabling diffusion-generated latent samples to remain aligned with physically meaningful porous structures during inverse-design refinement.

\section{Conclusion}

This paper presented a physical-property-guided latent diffusion framework for controllable 3D porous-media generation and inverse design. The framework combines property-aware latent modelling, conditional latent diffusion, differentiable structure-to-property feedback, and Palabos-based verification. By jointly refining the diffusion denoiser and pVAE decoder, the method reduces the mismatch between diffusion-generated latent samples and the decoder latent distribution, improving target-property consistency without requiring expensive online simulation.

Experiments on both procedurally generated porous structures and real micro-CT porous-media datasets showed that the proposed method improves simulator-verified inverse-design performance compared with representative pVAE-based and latent-diffusion-based baselines. The results demonstrate lower overall target-matching scores, stronger permeability controllability, and higher correlations between target and Palabos-verified properties. They also highlight that visual realism alone is insufficient for porous-media inverse design, as visually plausible structures may still deviate substantially from desired transport properties.

Future work will extend the framework to additional physical quantities such as tortuosity, reactive surface area, thermal transport, and multiphase flow behaviour.

\bibliographystyle{IEEEtran}
\bibliography{inversepore}

@article{kench2022microlib,
  title={MicroLib: A library of 3D microstructures generated from 2D micrographs using SliceGAN},
  author={Kench, Steve and Squires, Isaac and Dahari, Amir and Cooper, Samuel J},
  journal={Scientific Data},
  volume={9},
  number={1},
  pages={645},
  year={2022},
  publisher={Nature Publishing Group UK London}
}

@article{volkhonskiy2022generative,
  title={Generative adversarial networks for reconstruction of three-dimensional porous media from two-dimensional slices},
  author={Volkhonskiy, Denis and Muravleva, Ekaterina and Sudakov, Oleg and Orlov, Denis and Burnaev, Evgeny and Koroteev, Dmitry and Belozerov, Boris and Krutko, Vladislav},
  journal={Physical Review E},
  volume={105},
  number={2},
  pages={025304},
  year={2022},
  publisher={APS}
}

@article{kononov2023reconstruction,
  title={Reconstruction of 3d random media from 2d images: generative adversarial learning approach},
  author={Kononov, Evgeniy and Tashkinov, Mikhail and Silberschmidt, Vadim V},
  journal={Computer-Aided Design},
  volume={158},
  pages={103498},
  year={2023},
  publisher={Elsevier}
}

@article{ren2024using,
  title={Using Physics Informed Generative Adversarial Networks to Model 3D porous media},
  author={Ren, Zihan and Srinivasan, Sanjay},
  journal={arXiv preprint arXiv:2409.11541},
  year={2024}
}

@article{dureth2023conditional,
  title={Conditional diffusion-based microstructure reconstruction},
  author={D{\"u}reth, Christian and Seibert, Paul and R{\"u}cker, Dennis and Handford, Stephanie and K{\"a}stner, Markus and Gude, Maik},
  journal={Materials Today Communications},
  volume={35},
  pages={105608},
  year={2023},
  publisher={Elsevier}
}

@article{naiff2025controlled,
  title={Controlled Latent Diffusion Models for 3D Porous Media Reconstruction},
  author={Naiff, Danilo and Schaeffer, Bernardo P and Pires, Gustavo and Stojkovic, Dragan and Rapstine, Thomas and Ramos, Fabio},
  journal={arXiv preprint arXiv:2503.24083},
  year={2025}
}

@article{zhu2025diffusion,
  title={Diffusion Model-Based Generation of Three-Dimensional Multiphase Pore-Scale Images},
  author={Zhu, Linqi and Bijeljic, Branko and Blunt, Martin J},
  journal={Transport in Porous Media},
  volume={152},
  number={3},
  pages={22},
  year={2025},
  publisher={Springer}
}

@article{amiri2024new,
  title={New 2d to 3d reconstruction of heterogeneous porous media via deep generative adversarial networks (GANs)},
  author={Amiri, Hamed and Vogel, Hannah and Pl{\"u}mper, Oliver},
  journal={Journal of Geophysical Research: Machine Learning and Computation},
  volume={1},
  number={3},
  pages={e2024JH000178},
  year={2024},
  publisher={Wiley Online Library}
}

@article{zhang2024vegan,
  title={DA-VEGAN: Differentiably Augmenting VAE-GAN for microstructure reconstruction from extremely small data sets},
  author={Zhang, Yichi and Seibert, Paul and Otto, Alexandra and Ra{\ss}loff, Alexander and Ambati, Marreddy and K{\"a}stner, Markus},
  journal={Computational Materials Science},
  volume={232},
  pages={112661},
  year={2024},
  publisher={Elsevier}
}

@article{wang2020deep,
  title={Deep generative modeling for mechanistic-based learning and design of metamaterial systems},
  author={Wang, Liwei and Chan, Yu-Chin and Ahmed, Faez and Liu, Zhao and Zhu, Ping and Chen, Wei},
  journal={Computer Methods in Applied Mechanics and Engineering},
  volume={372},
  pages={113377},
  year={2020},
  publisher={Elsevier}
}

@article{nguyen2026deep,
  title={Deep learning-aided inverse design of porous metamaterials},
  author={Nguyen, Phu Thien and Heider, Yousef and Kochmann, Dennis M and Aldakheel, Fadi},
  journal={Computer Methods in Applied Mechanics and Engineering},
  volume={449},
  number={11849},
  pages={118499},
  year={2026},
  publisher={Elsevier}
}

@article{phu2024investigating,
  title={Investigating the impact of deformation on foam permeability through CT scans and the Lattice-Boltzmann method},
  author={Phu, Nguyen Thien and Navrath, Uwe and Heider, Yousef and Carmai, Julaluk and Markert, Bernd},
  journal={PAMM},
  volume={24},
  number={1},
  pages={e202300154},
  year={2024},
  publisher={Wiley Online Library}
}

@article{jones2024multiscale,
  title={Multiscale simulation of spatially correlated microstructure via a latent space representation},
  author={Jones, Reese E and Hamel, Craig M and Bolintineanu, Dan and Johnson, Kyle and de Macedo, Robert Buarque and Fuhg, Jan and Bouklas, Nikolaos and Kramer, Sharlotte},
  journal={International Journal of Solids and Structures},
  volume={301},
  pages={112966},
  year={2024},
  publisher={Elsevier}
}

@article{alzahrani2023pore,
  title={Pore-GNN: A graph neural network-based framework for predicting flow properties of porous media from micro-CT images},
  author={Alzahrani, Mohammed K and Shapoval, Artur and Chen, Zhixi and Rahman, Sheikh S},
  journal={Advances in Geo-Energy Research},
  volume={10},
  number={1},
  pages={39--55},
  year={2023}
}

@article{nguyen2022synthesizing,
  title={Synthesizing controlled microstructures of porous media using generative adversarial networks and reinforcement learning},
  author={Nguyen, Phong CH and Vlassis, Nikolaos N and Bahmani, Bahador and Sun, WaiChing and Udaykumar, HS and Baek, Stephen S},
  journal={Scientific reports},
  volume={12},
  number={1},
  pages={9034},
  year={2022},
  publisher={Nature Publishing Group UK London}
}

@article{lavigne2025synthetic,
  title={Synthetic Porous Microstructures: Automatic Design, Simulation, and Permeability Analysis},
  author={Lavigne, Thomas and Afanador, Camilo Andr{\'e}s Suarez and Obeidat, Anas and Urcun, St{\'e}phane},
  journal={arXiv preprint arXiv:2502.14518},
  year={2025}
}

@inproceedings{rombach2022high,
  title={High-resolution image synthesis with latent diffusion models},
  author={Rombach, Robin and Blattmann, Andreas and Lorenz, Dominik and Esser, Patrick and Ommer, Bj{\"o}rn},
  booktitle={Proceedings of the IEEE/CVF conference on computer vision and pattern recognition},
  pages={10684--10695},
  year={2022}
}

@article{zhou2022two,
  title={Two-order deep learning for generalized synthesis of radiation patterns for antenna arrays},
  author={Zhou, Zhao and Wei, Zhaohui and Ren, Jian and Yin, Yingzeng and Pedersen, Gert Fr{\o}lund and Shen, Ming},
  journal={IEEE Transactions on Artificial Intelligence},
  volume={4},
  number={5},
  pages={1359--1368},
  year={2022},
  publisher={IEEE}
}

@article{nguyen2025camox,
  title={CamoX: A Diffusion-based Method with Few-shot Learning for Environment-guided Camouflage Pattern Generation},
  author={Nguyen, Tran Thanh Phong and Gulrez, Tauseef and Culpepper, Joanne B and Phung, Son Lam and Le, Hoang Thanh},
  journal={IEEE Transactions on Artificial Intelligence},
  year={2025},
  publisher={IEEE}
}

@article{klopries2025itf,
  title={ITF-VAE: Variational Auto-Encoder using interpretable continuous time series features},
  author={Klopries, Hendrik and Schwung, Andreas},
  journal={IEEE Transactions on Artificial Intelligence},
  year={2025},
  publisher={IEEE}
}
\end{document}